\documentclass[10pt,twocolumn,letterpaper]{article}

\usepackage[pagenumbers]{cvpr} 

\usepackage{multirow}
\usepackage{amsmath}
\usepackage{xfrac}
\usepackage[accsupp]{axessibility} 
\usepackage{colortbl}

\usepackage[dvipsnames]{xcolor}

\definecolor{cvprblue}{rgb}{0.21,0.49,0.74}
\usepackage[pagebackref,breaklinks,colorlinks,citecolor=cvprblue]{hyperref}
\usepackage{tikz}
\usepackage{graphicx}
\usepackage{wrapfig}
\usepackage{makecell}
\usepackage{adjustbox}
\usepackage{hhline}

\usepackage{algorithm}
\usepackage{bm}
\usepackage{algorithmic}
\usepackage{listings}

\usepackage{pifont} 
\newcommand{\method}{{CrowdDiff}}

\usepackage[hyphens]{url} 
\usetikzlibrary{positioning}

\usepackage{mathtools}
\usepackage{siunitx}
\newcommand{\sota}{state-of-the-art}
\newcommand{\snr}{SNR}
\newcommand{\mapfont}[1]{\fontsize{#1}{#1}\selectfont}

\newcommand{\qnrf}{UCF-QNRF}
\newcommand{\shha}{ShanghaiTech A}
\newcommand{\ucf}{UCF-CC-50}
\newcommand{\shhb}{ShanghaiTech B}
\newcommand{\jhu}{JHU-Crowd++}
\newcommand{\nwpu}{NWPU-Crowd}

\newcommand{\tab}[1]{Table {#1}}

\newcommand{\first}[1]{\textbf{\textcolor{red}{#1}}}
\newcommand{\second}[1]{\textbf{\textcolor{blue}{#1}}}

\def\tablegap{-3 mm}
\title{\textit{CrowdDiff}: Multi-hypothesis Crowd Density Estimation using Diffusion Models}

\author{Yasiru Ranasinghe, Nithin Gopalakrishnan Nair, Wele Gedara Chaminda Bandara, and Vishal M. Patel\\
Johns Hopkins University, Baltimore, USA\\
{\tt\small \{dranasi1, ngopala2, wbandar1, vpatel36\}@jhu.edu}
}

\begin{document}
\maketitle

\setlength{\belowdisplayskip}{0pt} \setlength{\belowdisplayshortskip}{0pt}
\setlength{\abovedisplayskip}{0pt} \setlength{\abovedisplayshortskip}{0pt}

\begin{abstract}
Crowd counting is a fundamental problem in crowd analysis which is typically accomplished by estimating a crowd density map and summing over the density values. However, this approach suffers from background noise accumulation and loss of density due to the use of broad Gaussian kernels to create the ground truth density maps. This issue can be overcome by narrowing the Gaussian kernel. However, existing approaches perform poorly when trained with ground truth density maps with broad kernels. To deal with this limitation, we propose using conditional diffusion models to predict density maps, as diffusion models show high fidelity to training data during generation. With that, we present \method\ that generates the crowd density map as a reverse diffusion process. Furthermore, as the intermediate time steps of the diffusion process are noisy, we incorporate a regression branch for direct crowd estimation only during training to improve the feature learning. In addition, owing to the stochastic nature of the diffusion model, we introduce producing multiple density maps to improve the counting performance contrary to the existing crowd counting pipelines. We conduct extensive experiments on publicly available datasets to validate the effectiveness of our method. \method\ outperforms existing \sota\ crowd counting methods on several public crowd analysis benchmarks with significant improvements. \method\ project is available at: \href{https://dylran.github.io/crowddiff.github.io}{https://dylran.github.io/crowddiff.github.io}.
\end{abstract}

\begin{figure}
    \centering
    \begin{tikzpicture}
    
    \node (img1) 
    {\includegraphics[width=4.0cm,height=2.7cm]{figures/intro_figure/crowd_image.jpg}}; 
    \node[below=of img1, anchor=center, yshift=0.85cm, ] {(a) Ground truth: 1155 (4)};
    
    \node[right=of img1,xshift=-1.2cm] (img2) 
    {\includegraphics[width=4.0cm,height=2.7cm]{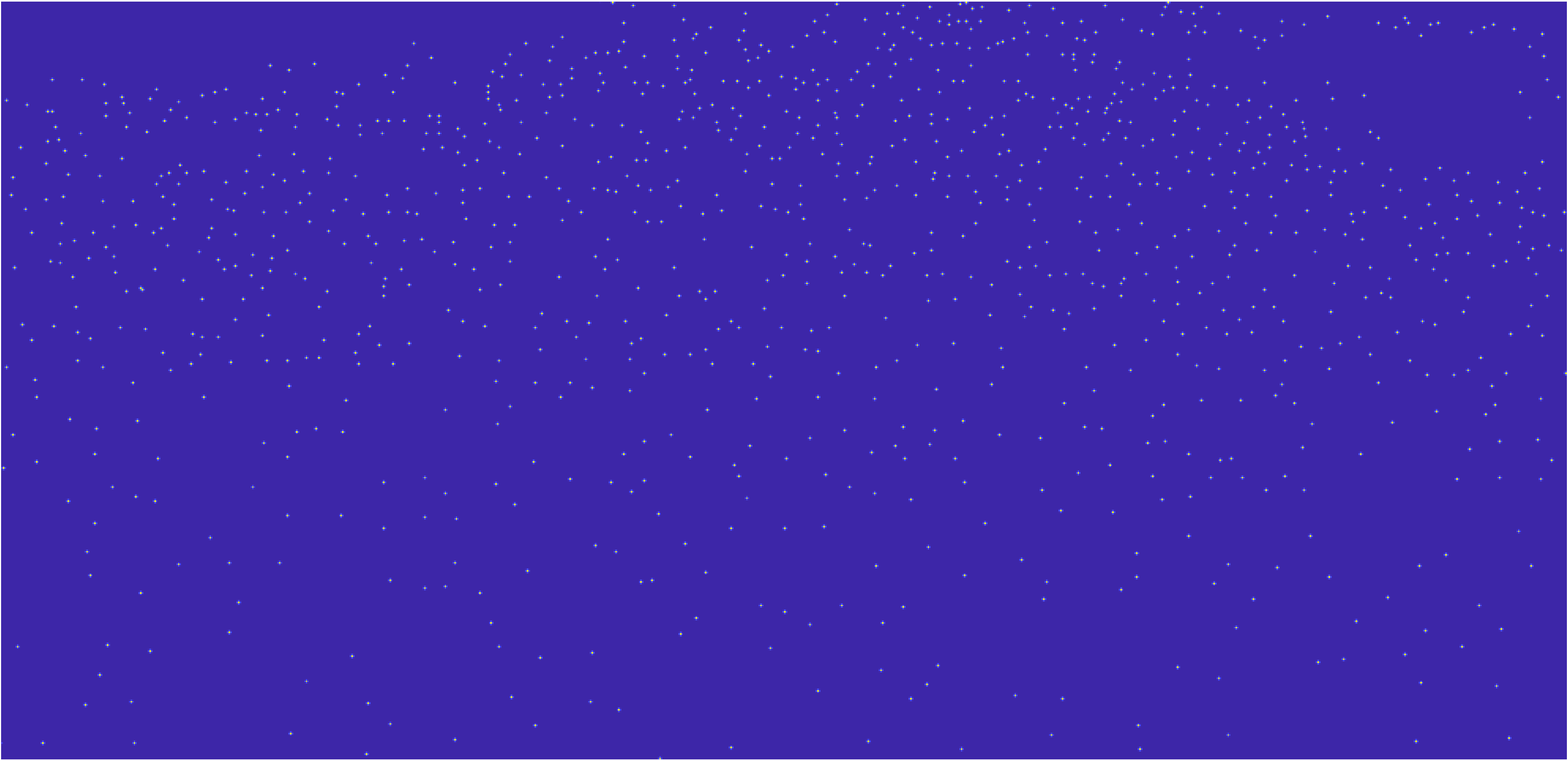}};
    \node[below=of img2, anchor=center, yshift=0.85cm, ] {(b) Ours: 1142 (4)};

    \node[below=of img1, yshift=0.75cm] (img3) 
    {\includegraphics[width=4.0cm,height=2.7cm]{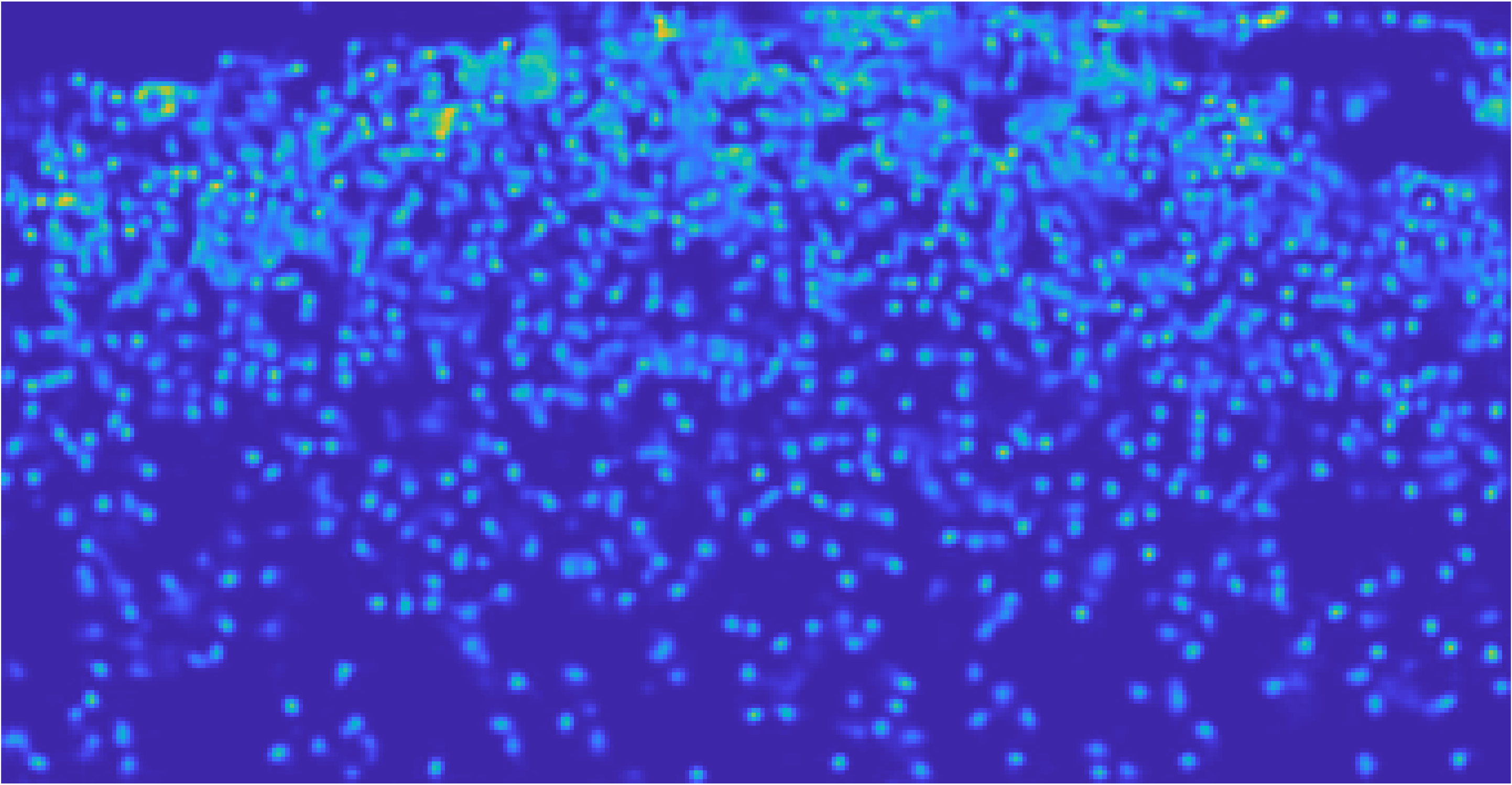}};
    \node[below=of img3, anchor=center, yshift=0.85cm, ] {(c) Chfl: 1187.6 (2.42)};

    \node[right=of img3,xshift=-1.2cm] (img4) 
    {\includegraphics[width=4.0cm,height=2.7cm]{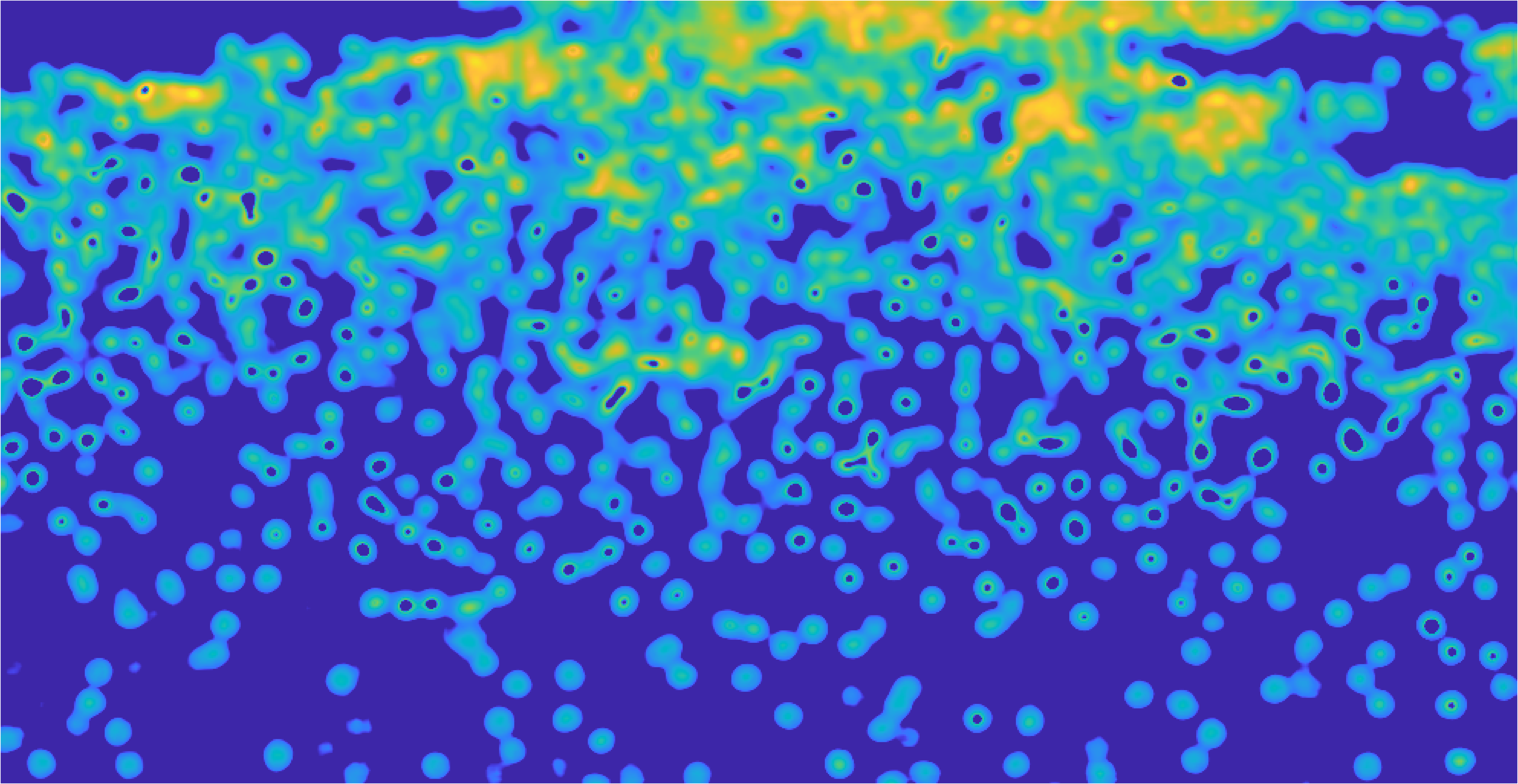}};
    \node[below=of img4, anchor=center, yshift=0.85cm, ] {(d) SUA: 1199.4 (3.20)};

    \node[right=of img1,xshift=-3.573cm,yshift=.957cm] (frame1) 
    {\includegraphics[height=0.15cm]{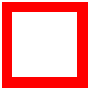}};
    \node[right=of img2,xshift=-3.573cm,yshift=.957cm] (frame2) 
    {\includegraphics[height=0.15cm]{figures/intro_crops/box.pdf}};
    \node[right=of img3,xshift=-3.573cm,yshift=.957cm] (frame3) 
    {\includegraphics[height=0.15cm]{figures/intro_crops/box.pdf}};
    \node[right=of img4,xshift=-3.573cm,yshift=.957cm] (frame4) 
    {\includegraphics[height=0.15cm]{figures/intro_crops/box.pdf}};

    \node[right=of img1,xshift=-2.235cm,yshift=-0.865cm] (frame1) 
    {\includegraphics[height=1.0cm]{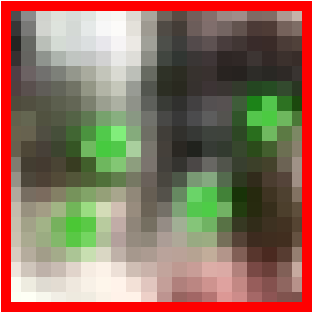}};
    \node[right=of img2,xshift=-2.235cm,yshift=-0.865cm] (frame1) 
    {\includegraphics[height=1.0cm]{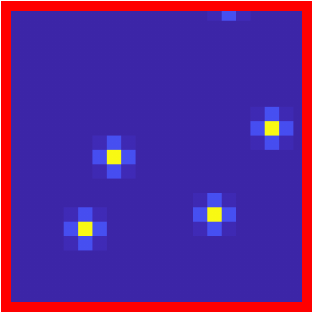}};
    \node[right=of img3,xshift=-2.235cm,yshift=-0.865cm] (frame1) 
    {\includegraphics[height=1.0cm]{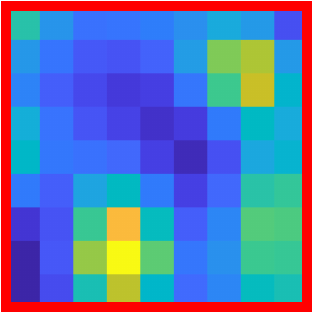}};
    \node[right=of img4,xshift=-2.235cm,yshift=-0.865cm] (frame1) 
    {\includegraphics[height=1.0cm]{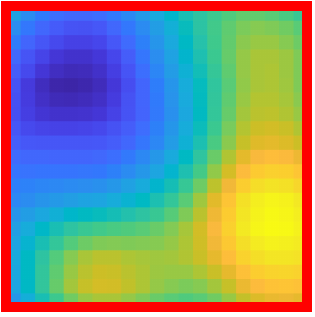}};
    
\end{tikzpicture}
    \caption{Predicted density results for (a) a dense crowd from (b) our method, (c) Chfl \cite{shu2022crowd}, and (d) SUA \cite{meng2021spatial}. The count of the enlarged crop is given in brackets.}
    \label{figure: density comparison}
\end{figure}

\section{Introduction}
\label{sec: introduction}
Crowd counting has been a fundamental problem in surveillance, public safety, and crowd control. Various methods have been proposed in the literature, including methods that directly predict the count \cite{liang2022transcrowd, zhang2019attentional, wang2019learning} or use a surrogate task such as density estimation \cite{hu2020count, jiang2020attention, wang2020distribution, song2021choose, wan2020kernel, tian2021cctrans}, object detection \cite{liu2019point, sam2020locate}, or point localization \cite{song2021rethinking, liang2022end, xu2022autoscale, gao2023application}. 

While density-based methods sum the estimated pixel density values for counting \cite{hu2020count}, localization-based methods count proposals with confidence scores higher than a threshold \cite{song2021rethinking}. As a result, density-based methods are more susceptible to introducing background noise into the final count compared to localization-based methods \cite{wan2019adaptive}. Furthermore, density estimation methods are affected by variations in crowd density distributions that arise due to different congestion levels of the crowd \cite{bai2020adaptive}. This could result in a loss of accuracy in the density estimation. In contrast, recent localization-based methods with point queries do not have the issue of background noise accumulation \cite{liang2022end}, like in density-based methods \cite{shu2022crowd}, as there is no interference between neighboring point proposals. However, localization-based methods require crowd density heuristics for proposal setting \cite{song2021rethinking}, which is not required by density-based methods. Thus, if the premise of point supervision is translated into density-based methods, it is possible to circumvent the requirement for crowd density heuristics, and the flaws of conventional density-based methods, and a narrow density kernel can be used to achieve this. However, Xu et al. \cite{xu2022autoscale} demonstrated that using a narrow kernel with density regression methods is ineffective.

Alternatively, it is feasible to use a generative model to predict the density map of a given crowd image that would learn the distribution of the values in the density map. Though Generative Adversarial Network (GAN) based architectures have been used for density map prediction \cite{yao2020mask, duan2021mask, shen2018crowd}, these methods still rely upon broad kernel sizes and overlook the benefits of point supervision. Since the model learns the distribution of the density pixel values, it is advantageous to maintain the sample space of the density pixel values, and employing a broad kernel will only discourage it. Furthermore, the use of both point supervision and crowd density prediction with generative models has not been thoroughly investigated before. Also, the aforementioned GAN-based approaches restrict to a single crowd density map realization similar to the regression-based methods and jettison the stochastic nature of generative models to produce multiple density map realizations, which could improve the counting performance.

We propose using denoising diffusion probabilistic models (diffusion models), \cite{ho2020denoising, nichol2021improved} to generate crowd density maps for a given image. Though diffusion models have been applied to segmentation \cite{amit2021segdiff, gu2022diffusioninst}, super-resolution \cite{li2022srdiff}, object detection \cite{chen2022diffusiondet}, etc., to the best of our knowledge, neither crowd counting nor density map generation has been studied with diffusion models. Furthermore, with the narrow kernel, we minimize the interference between adjacent densities, which helps to maintain the bounds and the distribution of density pixel values. This, in turn, simplifies the distribution learning for the diffusion model and improves density prediction as illustrated in \cref{figure: density comparison}, where the proposed method has reproduced the narrow kernel even in a dense region, while the other two recent methods failed.

Additionally, to eschew the probable loss of density with the density-based crowd counting methods, we count the number of blobs observed in the predicted density map by thresholding pixel density values. Consequently, we eliminate the effect of background noise as there is no requirement to sum over the density pixel values. Then, we introduce the crowd map fusion mechanism, combining multiple dot maps constructed after thresholding to improve the counting performance. This is only possible with generative models due to their stochastic nature. In addition, inspired by \cite{deja2023learning} on joint learning with diffusion models, we introduce an auxiliary regression branch only during training, which estimates the count based on encoder-decoder features  from the denoising network to improve feature learning. 

In summary, our contributions are:
\begin{itemize}[noitemsep]
    \item {\bf We formulate crowd density map generation as a denoising diffusion process.} \method\ is the first study to perform crowd counting with diffusion models.
    \item {\bf We promote using a narrow Gaussian kernel} to ease the learning process and facilitate the high-quality density map generation with more fidelity to the ground truth. 
    \item {\bf We propose a mechanism to consolidate multiple crowd density realizations}  to improve  performance utilizing the stochastic nature of diffusion models.
    \item {\bf We show that the proposed method surpasses the \sota\ performance} on public datasets.
\end{itemize}

\section{Background and Related Work}
\label{sec: related work}

\subsection{Crowd counting}
\label{subsec: crowd counting}
\textbf{Localization-based methods} perform counting by predicting the locations of heads, and generally, they involve predicting a bounding box \cite{lian2019density, liu2019point, sam2020locate, zhong2022mask} for each head. The literature has also proposed localization by points \cite{lian2019density} or blobs \cite{liu2019context}.  Recently, to remove the necessity of post-processing, such as non-maximum suppression, point localization \cite{song2021rethinking, liang2022end} was introduced to crowd counting.\\
\textbf{Density-based methods} \cite{liu2019counting, liu2020weighing, xiong2019open, li2018csrnet, miao2020shallow, liu2020adaptive, bai2020adaptive} attempt to produce a density map for a given crowd image. However, density-based methods suffer from background noise and loss of density \cite{ma2021learning, oh2020crowd, luo2020hybrid} in congested regions due to broad kernels. But, using a narrow Gaussian kernel to generate ground truths is ineffective according to \cite{xu2022autoscale} with regression networks. Hence, we treat the prediction of the density map as a generative task.

\subsection{Diffusion models for crowd density generation}
\label{subsec: diffusion model}
Diffusion models \cite{sohl2015deep} are defined based on a Markov chain with a forward and a reverse process. In the forward process, noise is gradually added to data; and is denoised in the reverse process. The forward process is formulated as,
    \[q(\mathbf{x_t}|\mathbf{x_{t-1}}) = \mathcal{N}(\mathbf{x_t}|\sqrt{1-\beta_t}\mathbf{x_{t-1}},\beta_t\mathbf{I}),\]
where the sample datum $\mathbf{x_0}$ is gradually transformed to a noisy sample $\mathbf{x_t}$ for $t \in \{1,\dots,T\}$ by adding Gaussian noise according to a noise variance schedule $\beta_1,\dots,\beta_T$. Here, $\mathbf{I}$ is the identity matrix. Nonetheless, $\mathbf{x_t}$ can be computed using $\mathbf{x_0}$ and a noise vector $\epsilon \sim \mathcal{N}(\mathbf{0},\mathbf{I})$ and with the forward transformation,
    \[\mathbf{x_t} = \sqrt{\bar{\alpha_t}}\mathbf{x_0} + \sqrt{\left(1-\bar{\alpha_t} \right)}\epsilon,\]
where $\bar{\alpha_t} \coloneqq \prod_{\tau=1}^t \alpha_\tau = \prod_{\tau=1}^t (1-\beta_\tau)$ and $\beta_\tau$.

In this work, we aim to perform crowd density map generation via the diffusion model. Hence, our data samples will be crowd density maps $\mathbf{x_0} \in \mathbb{R}^{H\times W}$, where $H \text{ and } W$ are the height and width dimensions. However, in lieu of training a neural network to predict $\mathbf{x_0}$ from $\mathbf{x_t}$ for various time steps, we predict the amount of noise ($\hat{\epsilon}$) in $\mathbf{x_t}$ at each time step conditioned on the crowd image ($\mathbf{y}$) and apply the reverse diffusion process to obtain $\mathbf{x_0}$ ultimately. 

To that end, to train the denoising diffusion network, we use the hybrid loss ($\mathcal{L}_{hybrid}$) function proposed in \cite{nichol2021improved}. To promote learning coarse features at lower \snr\ stages, we adopt the weighting scheme \cite{choi2022perception} defined as,
\begin{equation}
\label{equation: lambda}
    \lambda_t = \frac{\sfrac{(1-\beta_t)(1-\bar{\alpha_t})}{\beta_t}}{{\left(k + \text{\snr}{\scriptstyle(t)}\right)}^\gamma}, \text{where}\;\; \text{\snr}{\scriptstyle(t)} = \frac{\bar{\alpha_t}}{1-\bar{\alpha_t}}
\end{equation}
with $k$ and $\gamma$ as hyperparameters. Hence, the final loss over which the denoising network is optimized is  as follows,
    \[\mathcal{L}_{hybrid} = \mathbf{E}_{\mathbf{x_0},\mathbf{y},\epsilon} \left[ \lambda_t{\lVert\hat{\epsilon}_{(\mathbf{x_0},\mathbf{y},t)} - \epsilon\rVert}_2^2\right] + \lambda_{vlb}~\mathcal{L}_{vlb},\]
where $\mathcal{L}_{vlb}$ is the original variational lower bound defined in \cite{nichol2021improved} and $\lambda_{vlb}$ is its weighting factor.

\begin{figure}[!t]
\begin{center}
   \includegraphics[width=\linewidth]{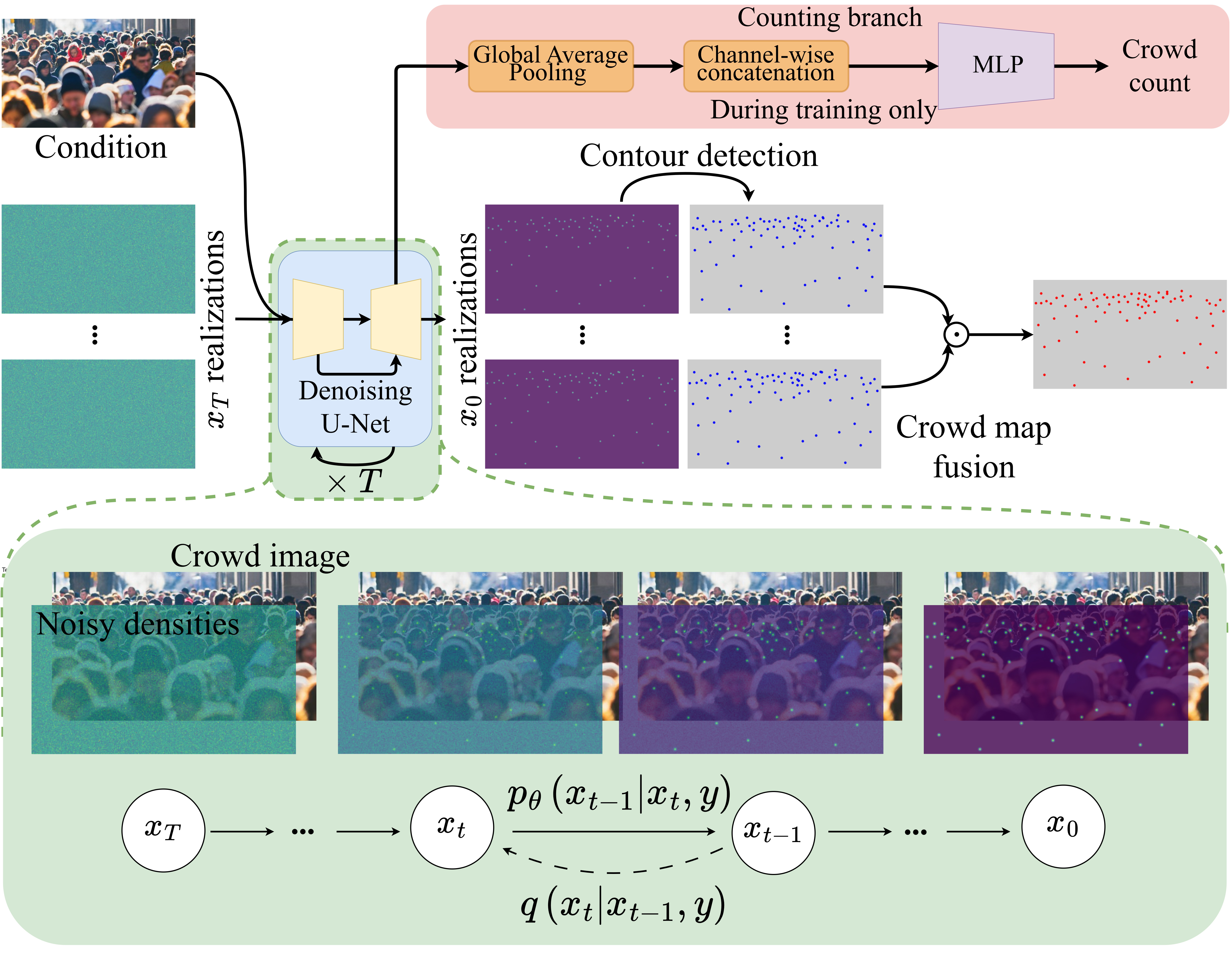}
\end{center}
    \caption{Overall crowd counting pipeline. The crowd density maps are generated from the denoising diffusion process for a crowd image. Next, thresholding is performed on the resulting crowd density realizations to create crowd maps. The crowd maps are then fused into a single crowd map. The counting branch is trained in parallel using the encoder-decoder features of the denoising U-Net and discarded during inference.}
\label{figure: diffusion crowd count pipeline}

\end{figure}

\section{\method}
\label{sec: methods}
In this section, we first review the motivation for selecting an appropriate kernel size. We present the joint learning of counting as an auxiliary task to improve the density map generation performance. Finally, we introduce a method to combine different realizations for density maps to improve crowd counting performance. The overall crowd counting pipeline is illustrated in \cref{figure: diffusion crowd count pipeline}.

\subsection{Narrow kernel}
\label{subsec: narrow kernel}
The diffusion process requires a density map to learn the conditional distribution of crowd density. The crowd density map can be acquired by convolving point information with a pre-defined Gaussian kernel. For that, selecting a proper kernel size and variance is important as it governs the distribution of the pixel values of the crowd density maps.

\begin{figure}[!t]
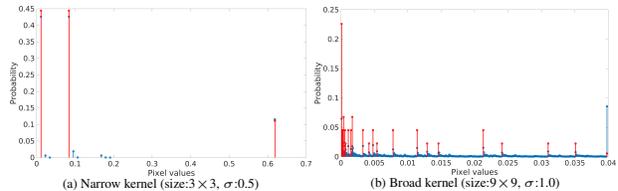

    \centering
\begin{tikzpicture}
    \node (img1)
    {\includegraphics[width=4.0cm]{figures/pixel_density_distribution/narrow.png}};
    \node[below=of img1, anchor=center, yshift=1.0cm, font=\mapfont{5}] {(a) Narrow kernel (size:3$\times$3, $\sigma$:0.5)};
    \node[right=of img1,xshift=-1.2cm] (img2) 
    {\includegraphics[width=4.0cm]{figures/pixel_density_distribution/sparse.png}};
    \node[below=of img2, anchor=center, yshift=1.0cm, font=\mapfont{5}] {(b) Broad kernel (size:9$\times$9, $\sigma$:1.0)};    
\end{tikzpicture}

    \caption{Change in the pixel values of the Gaussian kernel (red stems) and the resulting density map (blue stems) for a crowd image with a $3,547$ crowd count. The kernel size and variance increase from left to right.}
    
    \label{figure: pixel value distribution}
\end{figure}

As demonstrated in \cref{figure: pixel value distribution}, the divergence between the distribution of the Gaussian kernel (values) and the resulting density map increases as the kernel size and variance increase, especially for the congested scene. This might not be the case for sparse crowd scenes, as there is minimal or no interference between density kernels. However, this implies that the density pixel value distribution is highly image-dependent, hindering the crowd densities' learning. This can be eschewed by narrowing the distribution of the Gaussian kernel as illustrated in \cref{figure: pixel value distribution}. This also helps the denoising  network to maintain the pixel values within a pre-defined range. The difference between the probability mass of a broad Gaussian kernel and the resulting density map is significant. This can lead to the clipping of many pixel values, resulting in a loss of information in congested scenes.

The aforementioned issue can be solved by a narrow kernel. A narrow kernel provides an alternative path to crowd counting without summing over the density map values. As shown in \cref{figure: density comparison}, the crowd count can be obtained by simply counting the observable kernels. For that, we perform thresholding on the density maps and obtain the location of each kernel. Then, the crowd count is computed as the total number of locations. This provides the means to avoid background noise in the generated density maps and to obtain the crowd count by detecting these narrow kernels in the crowd density maps. Unlike the local maxima detection strategy proposed in \cite{liang2022focal} to detect head locations from a crowd density map, our method is not dependent on any hyperparameter tuning for detection.

\subsection{Joint learning of counting}
\label{subsec: joint learning}
Directly regressing the crowd count from image features is a difficult task \cite{liang2022transcrowd} compared to counting with a surrogate task. To perform the direct computation of the crowd count, we consider the intermediate features of the encoder-decoder of the denoising U-Net. Let's denote the set of intermediate features from the denoising network for a particular timestep $t$ as $\mathcal{Z}_t = \{\mathbf{z}^1_t,\mathbf{z}^2_t,\dots,\mathbf{z}^d_t\}$, where $\mathbf{z}^{*}_t$ is the representation vector at the corresponding feature level of the decoder. Since the spatial dimensions of the intermediate representations at different depth levels are incompatible, global average pooling is performed on each $\mathbf{z}^*_t$, which are then concatenated to construct a single feature vector $\mathbf{z}_t$. This is then passed through the regression network to estimate the crowd count at various noise levels.

However, for a sampled pair ($\mathbf{x_0}, \mathbf{y}$), only the density map $\mathbf{x_0}$ is diffused with noise according to a noise schedule. Hence, the noise level in the set of intermediate features $\mathcal{Z}_t$ will vary with the timestep, and \snr\ will be lower in the later stages of the diffusion process than in the earlier stages. Hence, we utilize the weighting scheme discussed in \cref{subsec: diffusion model} during the training of the count regression network. We utilize $\mathcal{L}_1$ loss as follows, 
    \[\mathcal{L}_1^t = \lambda_t{\lVert\bar{c_t} - c\rVert}_1\]
to measure the difference between the prediction ($\bar{c_t}$) and the ground truth ($c$) for a given time step $t$ and a given sampled pair, where $\lambda_t$ is the same weighting factor used in \cref{equation: lambda}. As the training loss for the denoising model is the Monte-Carlo approximation of the sum over all timesteps, the training loss can be written as
    \[\mathcal{L}_{count} = \mathbb{E}_{\mathbf{x_0}, \mathbf{y}, t}\left[ \lambda_t{\lVert\bar{c_t} - c\rVert}_1 \right].\]
The overall training includes optimization over the parameters of the denoising network and the regression branch. Hence, the overall training objective is as follows,
    \[\mathcal{L}_{overall} = \mathcal{L}_{hybrid} + \lambda_{count}\mathcal{L}_{count},\]
where $\lambda_{count}$ is a weightage on the counting task.

\begin{figure}[!t]
\begin{center}
   \includegraphics[width=0.8\linewidth]{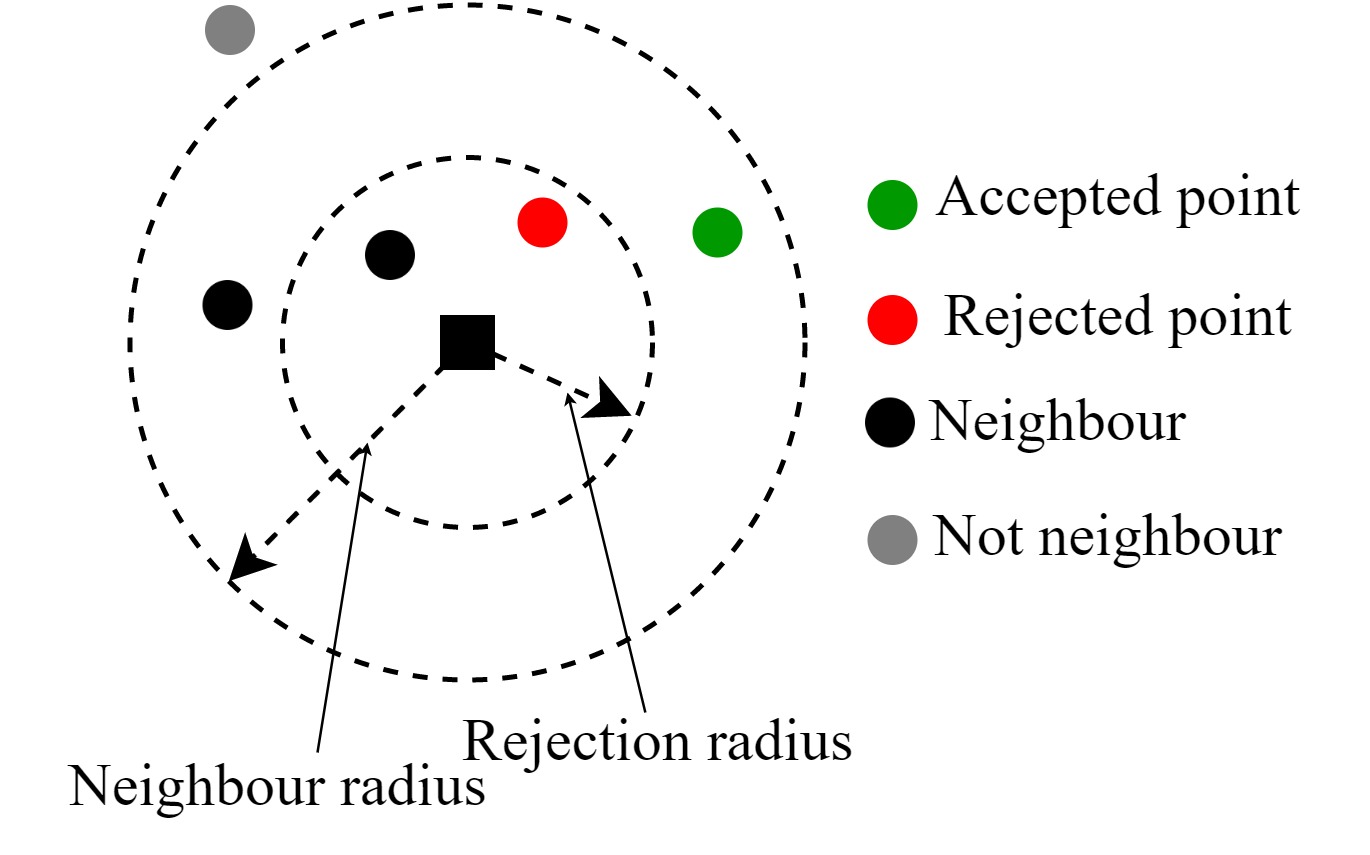}
   
\end{center}

    \caption{Crowd map fusion criterion. The rejection radius is computed from the neighbors (black) inside the neighbor radius. New points (colored) that fall inside the rejection radius are removed (red), and the rest (green) are combined into the compound map.}
    
\label{figure: density fusion method}
\end{figure}

\subsection{Stochastic crowd map fusion}
\label{ssec: crowd density fusion}
The stochastic nature of the diffusion models could generate different realizations of the crowd density map for the same crowd image. Therefore, the counting performance with diffusion models can be improved with multiple realizations contrary to the traditional crowd counting methods as evidenced by other tasks based on diffusion models such as segmentation \cite{gu2022diffusioninst} and detection \cite{chen2022diffusiondet}. However, rather than averaging individual counts from different realizations, they could be combined to compute a more improved count because individual realizations could infer crowd densities that were not present in other realizations.

To combine different realizations for the  density maps, only the new information should be transferred to the compound  density map. For that, we first compute the locations of the density kernels by density thresholding. Once these locations are found, a dot map is constructed for each density map, referred to as the `crowd map.' Then, we consider the dissimilarity between the crowd maps from different realizations, and to measure that, we consider the structural similarity index measure (SSIM) \cite{wang2004image}. We assign a similarity score measured as the cumulative SSIM with the remaining crowd map realizations for each crowd map. Then, the maps will be arranged in the ascending order of the SSIM before combining. Further, we don't require the ground truth locations to combine different realizations; they are combined depending on the similarity of the crowd maps. 

Let’s consider four crowd maps. For a given crowd map (source map), we’ll measure the SSIM with each of the remaining three crowd maps, and the sum of those three SSIMs will be assigned as the similarity score of the source map. If the similarity score is the highest of a map, then it is the most similar to the remaining maps and likely to contain most of the point locations available in the remaining maps. Hence, the new points that can be added to and from the most similar map are minimal. Conversely, the crowd map with the least similarity score differs the most from the remaining maps; therefore, the new points that can be added to\,/\,from this map are maximal. Consequently, the best map to start the fusion process is the crowd map with the lowest similarity score. Likewise, we order the crowd maps in the ascending order of the similarity score to combine.

When fusing two crowd maps, it is necessary to reject repeating point locations. This is performed based on the locations of the new points compared to the points in the combined list. We take the crowd map first and the head locations from that realization as the reference. 
Next, we define a rejection radius for each head location as:
    \[r_n = \beta \frac{\sum_{i=1}^{\Tilde{k}}r_{ni}}{2\Tilde{k}}\]
by considering the {\it k}-nearest neighbors within a fixed range. Here $\beta$ is a scaling factor and $\Tilde{k}$ is the total nearest neighbors within the range. Next, we remove the head locations of the next crowd map locations that fall within the rejection radii in the reference map as illustrated in \cref{figure: density fusion method}, and the remaining locations are added to the reference map. This procedure is performed until all realizations are exhausted.

\section{Experimental Details}
\label{sec: experimental details}
\textbf{\method\ pipelines.} \underline{During training}, we create the ground truth density map with narrow kernels as described in \cref{subsec: narrow kernel}. Next, we randomly sample a timestep $t$. Then, we sample a Gaussian noise according to the variance at $t$ and add it to the ground truth map, resulting in the noisy map ($\mathbf{x_t}$). Then, we input the image and $\mathbf{x_t}$ to the denoising U-Net (network) and predict the noise added to the ground truth. Hence, based on the crowd image, the network is trained to predict the noise in $\mathbf{x_t}$. \underline{During inference}, we sample a Gaussian noise from $\mathcal{N}\left(\mathbf{0,I}\right)$ at time $T$, which is used as the initial noisy density map $\mathbf{x_T}$. Then, the network will estimate the noise present in $\mathbf{x_T}$, and by removing that noise, we produce the noisy density map ($\mathbf{x_{T-1}}$) at time $T-1$. Likewise, we’ll repeat the process where the noisy density map $\mathbf{x_{t-1}}$ at $t-1$ is estimated from the noisy density map $\mathbf{x_t}$ at time $t$ until we produce the density map ($\mathbf{x_0}$) for the image. Besides that, the counting branch output is discarded during inference.\\
\textbf{Diffusion process} uses 1,000 timesteps and DDIM sampling \cite{song2020denoising} during inference. We use a linear noise schedule with noise variance ranging from \num{1e-3} to 0.02.\\
\textbf{Hyperparameter values} $\lambda_{count}$ is set as \num{5e-3} to match the range of the value for $L_{hybrid}$. The $\gamma$ and $k$ values are set as 0.5 and 1, respectively, to compute the \snr-based weighting factors. We adopt the original scaling factor of \num{1e-3} for $\lambda_{vlb}$ following \cite{nichol2021improved}. For crowd map fusion, we set $\beta$ equal to 0.85 and the maximum nearest neighbors to four. The radius for the neighbor search was restricted to 0.05 of the minimum of the image dimensions.\\
\textbf{Training} of the denoising network is initialized with the ImageNet pre-trained weights for the super-resolution \cite{rombach2022high} task except for the input and output layers. The network is trained for \num{2e5} iterations with a batch size of 8 for $256\times256$ images. We use an AdamW optimizer with a fixed learning rate \num{1e-4} and a linear warm-up schedule over \num{5e3} training steps following \cite{wu2023boosting}.\\
\textbf{Evaluations} are performed on six public datasets:  \jhu \cite{sindagi2019pushing}, \shha \cite{zhang2016single}, \shhb \cite{zhang2016single}, \ucf \cite{idrees2013multi}, \qnrf \cite{idrees2018composition}, and \nwpu\cite{wang2020nwpu}. We use MAE and MSE as the performance metrics.

\begin{table*}
\begin{center}
\caption{Comparison with \sota\ methods on the public crowd analysis benchmarks: \jhu, ShanghaiTech, UCF, and \nwpu. 
The best results are shown in \first{red}. The second-best results are shown in \second{blue}. 
}
\vspace{\tablegap}
\resizebox{0.95\textwidth}{!}{
\begin{tabular}{l c c c c c c c c c c c c c}
\toprule
 \multirow{2}{*}{Method} & \multirow{2}{*}{Venue} &\multicolumn{2}{c}{\jhu} &\multicolumn{2}{c}{\shha} &\multicolumn{2}{c}{\shhb} &\multicolumn{2}{c}{\ucf} &\multicolumn{2}{c}{\qnrf} &\multicolumn{2}{c}{\nwpu}\\[0.2ex]
 \cmidrule(lr){3-4}\cmidrule(lr){5-6}\cmidrule(lr){7-8}\cmidrule(lr){9-10}\cmidrule(lr){11-12}\cmidrule(lr){13-14}
& & MAE$\downarrow$ & MSE$\downarrow$ & MAE$\downarrow$ & MSE$\downarrow$ & MAE$\downarrow$ & MSE$\downarrow$ & MAE$\downarrow$ & MSE$\downarrow$ & MAE$\downarrow$ & MSE$\downarrow$ & MAE$\downarrow$ & MSE$\downarrow$\\[0.2ex]
\midrule\midrule
TopoCount \cite{abousamra2021localization}	& AAAI'21	& {60.9}	& {267.4}	& {61.2}	& {104.6}	& {7.8}	& {13.7}	& {184.1}	& {258.3}	& {89.0}	& {159.0}	& {107.8}	& {438.5}	\\[0.2ex]
SUA \cite{meng2021spatial}	& ICCV'21	& {80.7}	& {290.8}	& {68.5}	& {121.9}	& {14.1}	& {20.6}	& {-}	& {-}	& {130.3}	& {226.3}	& {111.7}	& {443.2}	\\[0.2ex]
ChfL \cite{shu2022crowd}	& CVPR'22	& {57.0}	& {235.7}	& {57.5}	& {94.3}	& {6.9}	& {11.0}	& {-}	& {-}	& {80.3}	& {137.6}	& {76.8}	& {343.0}	\\[0.2ex]
MAN \cite{lin2022boosting}	& CVPR'22	& {53.4}	& \second{209.9}	& {56.8}	& {90.3}	& {-}	& {-}	& {-}	& {-}	& {77.3}	& {131.5}	& {76.5}	& {323.0}	\\[0.2ex]
GauNet \cite{cheng2022rethinking}	& CVPR'22	& {58.2}	& {245.1}	& {54.8}	& {89.1}	& {6.2}	& {9.9}	& {186.3}	& {256.5}	& {81.6}	& {153.7}	& {-}	& {-}	\\[0.2ex]
CLTR \cite{liang2022end}	& ECCV'22	& {59.5}	& {240.6}	& {56.9}	& {95.2}	& {6.5}	& {10.6}	& {-}	& {-}	& {85.8}	& {141.3}	& {74.3}	& {333.8}	\\[0.2ex]
CrwodHat \cite{wu2023boosting}	& CVPR'23	& \second{52.3}	& {211.8}	& {51.2}	& {81.9}	& \first{5.7}	& {9.4}	& {-}	& {-}	& {75.1}	& \second{126.7}	& {68.7}	& \second{296.9}	\\[0.2ex]
STEERER \cite{han2023steerer}	& ICCV'23	& {54.3}	& {238.3}	& {54.5}	& {86.9}	& {5.8}	& \second{8.5}	& {-}	& {-}	& {74.3}	& {128.3}	& \second{63.7}	& {309.8}	\\[0.2ex]
PET \cite{liu2023point}	& ICCV'23	& {58.5}	& {238.0}	& \second{49.3}	& \second{78.8}	& {6.2}	& {9.7}	& {-}	& {-}	& {79.5}	& {144.3}	& {74.4}	& {328.5}	\\[0.2ex]
\rowcolor{black!10}\method\	& 	& \first{47.3}	& \first{198.9}	& \first{47.4}	& \first{75.0}	& \first{5.7}	& \first{8.2}	& \first{160.8}	& \first{225.0}	& \first{68.9}	& \first{125.6}	& \first{57.8}	& \first{221.2}	\\[0.2ex]
\bottomrule
\end{tabular}
}
\vspace{\tablegap}
\label{table: crowd counting performance}
\end{center}
\end{table*}

\begin{table*}[ht]
	\centering
	\caption{Comparison with state-of-the-art methods on the \nwpu\ \emph{test} dataset with performance under different scene constraints and luminance conditions. The best results are shown in \first{red}. The second-best results are shown in \second{blue}.} 
\vspace{\tablegap}
	\resizebox{0.95\textwidth}{!}{
		\begin{tabular}{lcccccccccccccc}
			\toprule
			\multirow{2}{*}{Method} & \multirow{2}{*}{Venue} & \multicolumn{3}{c}{Overall} & \multicolumn{6}{c}{Scene Level (MAE$\downarrow$)} & \multicolumn{4}{c}{Luminance (MAE$\downarrow$)}\\[0.2ex]
            \cmidrule(lr){3-5} \cmidrule(lr){6-11} \cmidrule(lr){12-15}
			& & MAE$\downarrow$ & MSE$\downarrow$ & NAE$\downarrow$ & Avg. & S0 & S1 & S2 & S3 & S4 & Avg. & L0 & L1 & L2 \\[0.2ex]
			\midrule \midrule
			BL \cite{ma2019bayesian}    & ICCV'19 & 105.4 & 454.2 & 0.203 & 750.5 & 66.5 & 8.7 & 41.2 & 249.9 & 3386.4 & 154.7 & 293.4 & 102.7 & 68.0 \\[0.2ex]
			DM-Count \cite{wang2020distribution} & NeurIPS'20 & 88.4 & 388.6 & 0.169 & 498.0 & 146.7 & 7.6 & 31.2 & 228.7 & 2075.8 & 117.6 & 203.6 & 88.1 & 61.2 \\[0.2ex]
			UOT \cite{ma2021learning}   & AAAI'21 &  87.8 & 387.5 & 0.185 & 566.5 & 80.7 & 7.9 & 36.3 & 212.0 & 2495.4 & 127.2 &240.3 & 86.4 & 54.9 \\[0.2ex]
			P2PNet \cite{song2021rethinking} & ICCV'21 & 72.6  & 331.6 & 0.192 & 510.0 & 34.7 & 11.3 & 31.5 & \second{161.0} & 2311.6 & 107.8 & 203.8 & 69.6 & 50.1 \\[0.2ex]
            MAN \cite{lin2022boosting} & CVPR'22 & 76.5& 323.0 & 0.170 & 464.6  &43.3 &8.5& 35.3&190.9 &2044.9 & 102.2 & 180.1&77.1& 49.4\\[0.2ex]
		    Chfl \cite{shu2022crowd}& CVPR'22 & 76.8& 343.0 & 0.171 & 470.1     &56.7 &8.4 &32.1 & 195.1 &2058.0 & 113.9 & 217.7&74.5 &49.6 \\[0.2ex]
            CLTR \cite{liang2022end} &ECCV'22 &74.4 &333.8 &0.165 & 532.4 & \second{4.2} & 7.3 & 30.3 & 185.5 & 2434.8 & 106.0 & 197.1 & 73.5 & 47.3\\[0.2ex]
			STEERER \cite{han2023steerer}  &  ICCV'23 & \second{63.7}  & 309.8 & \second{0.133} & 410.6  &  48.2 & \second{6.0} & \first{25.8} &\first{158.3} & 1814.5 & \second{87.2} &  \second{155.7}& \second{63.3}& \second{42.5} \\[0.2ex]
            CrowdHat \cite{han2023steerer}  &  CVPR'23 & 68.7  & \second{296.9} & 0.182 & \second{371.7}  &  {5.3} & 6.9 & 37.8 & 183.3 & \second{1625.3} & 108.8 &  220.4 & 66.3 & \first{39.6} \\[0.2ex]
             \rowcolor{black!10}\method & - &  \first{57.8} &  \first{221.2} &  \first{0.120} & \first{305.3} &  \first{4.1} &  \first{4.9} &  \second{28.8} &  166.2 &  \first{1322.4} & \first{79.7} &  \first{131.8} &  \first{53.1} &  54.3\\[0.2ex]
\bottomrule
		\end{tabular}
	}
\vspace{\tablegap}
\label{tab: nwpu_counting}
\end{table*}

\section{Results}
\label{sec: results and ablation}

\subsection{Crowd counting performance}
\label{subsec: main results}
\begin{figure}[!t]
\centering
\begin{tikzpicture}
    \node (img1)
    {\includegraphics[height=2.5cm]{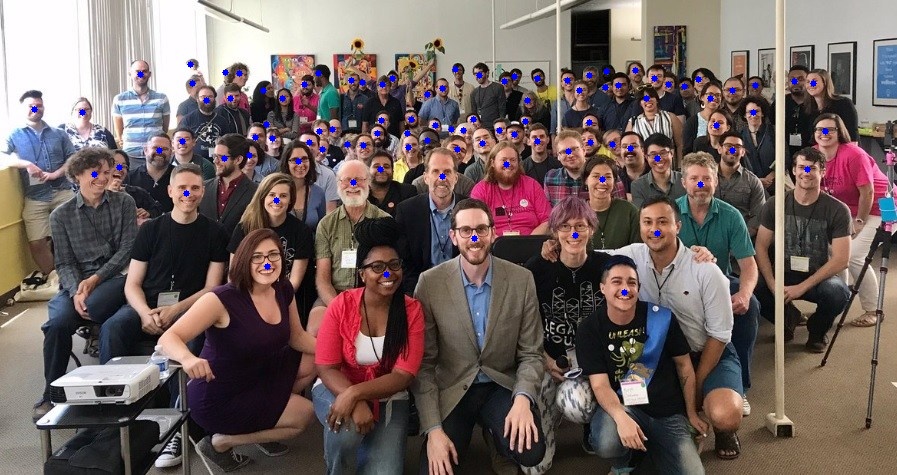}};
    \node[right=of img1,xshift=-1.2cm] (img4) 
    {\includegraphics[height=2.5cm]{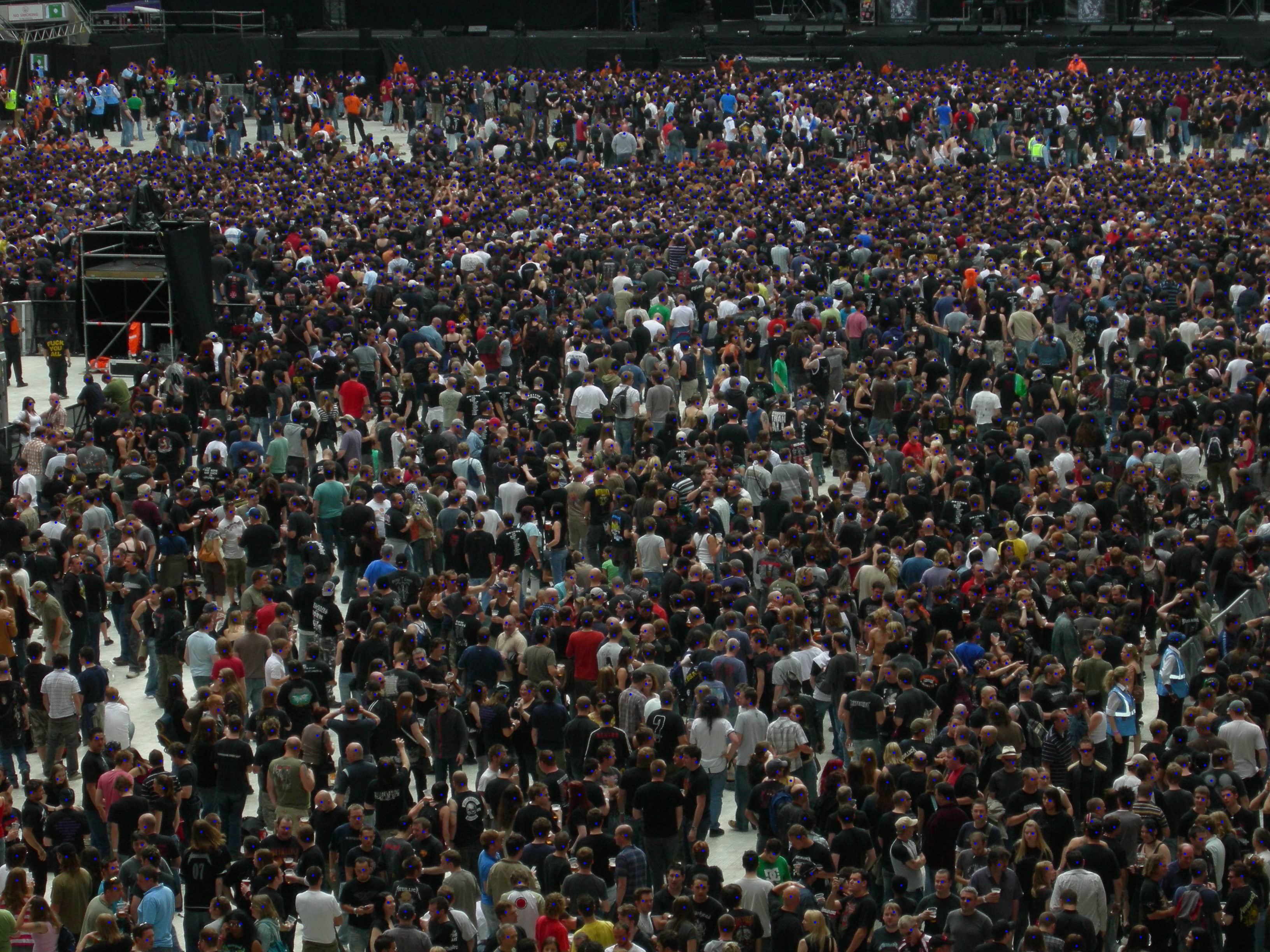}};
    
    \node[below=of img1, yshift=1.2cm] (img5) 
    {\includegraphics[height=2.5cm, width=4.75cm]{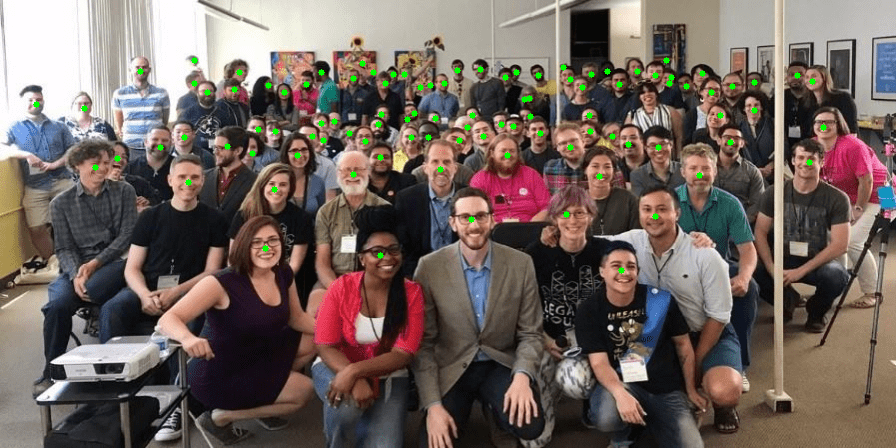}};
    \node[below=of img5, anchor=center, yshift=0.9cm, font=\mapfont{8}] {GT: 99     Prediction: 101};
    \node[right=of img5,xshift=-1.2cm] (img8) 
    {\includegraphics[height=2.5cm]{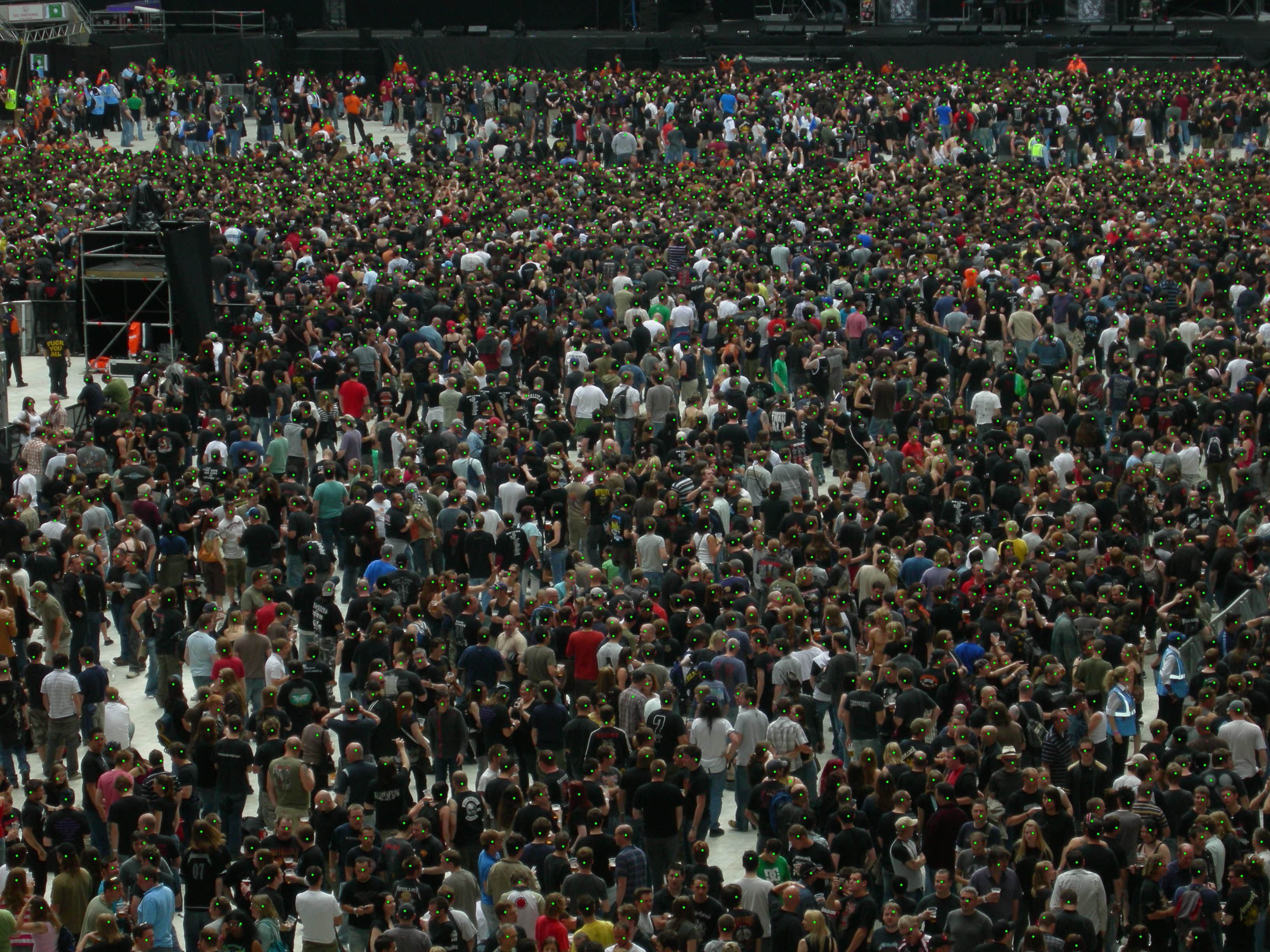}};
    \node[below=of img8, anchor=center, yshift=0.9cm, font=\mapfont{8}] {GT: 2607   Prediction: 2566};    
\end{tikzpicture}

\caption{Qualitative results for the proposed method with the ground truths. The prediction is produced after combining multiple realizations.}

\label{figure: crowd maps}
\end{figure}

\textbf{Quantitative results} for crowd counting are presented in \cref{table: crowd counting performance} for the proposed method with other existing methods. The proposed method achieves \sota\ crowd counting results on public crowd counting datasets, and two factors can explain the improvement. First, the proposed use of a narrow kernel has improved the counting results in dense regions by mitigating the loss of density values in contrast to conventional density-based methods. Second, we eliminate the effect of background noise on the crowd count, which scales with the image dimensions, by replacing density summation with thresholding followed by summation. The performance on \jhu, \qnrf, and \nwpu\ datasets explains the above effect of \method\ as these datasets contain dense crowd scenes and large image dimensions.
This was possible due to the capability to produce accurate density maps with better resemblance to ground truth maps with diffusion models. In \cref{tab: nwpu_counting}, we provide the performance on the \emph{test} set of the \nwpu\ dataset. In addition to the overall MAE, \method\ has the best performance in negative samples or sparse crowds similar to detection or localization-based methods. This is due to density thresholding because of the ability to produce narrow kernels without intersections.\\
\textbf{Qualitative results} are presented in \cref{figure: density comparison} and \cref{figure: crowd maps} for density map generation with diffusion models and crowd map generation with the proposed pipeline. As depicted in \cref{figure: density comparison}, the proposed method and the narrow kernel can accurately perform counting even in a dense region. In contrast, the other two methods have suffered from loss of density. Furthermore, our proposed pipeline has identified head locations accurately, which is impossible with existing density-based methods and without data heuristics, unlike localization-based methods.

\begin{figure}[!ht]
    \centering
\begin{tikzpicture}
    
    \node (img1) 
    {\includegraphics[width=4.0cm,height=2.7cm]{figures/GAN_ablation/IMG_104.jpg}}; 
    \node[below=of img1, anchor=center, yshift=0.85cm, ] {(a) Ground truth: 1174 };
    
    \node[right=of img1,xshift=-1.2cm] (img2) 
    {\includegraphics[width=4.0cm,height=2.7cm]{figures/GAN_ablation/IMG_94.jpg}};
    \node[below=of img2, anchor=center, yshift=0.85cm, ] {(b) Ground truth: 70 };

    \node[below=of img1, yshift=0.75cm] (img3) 
    {\includegraphics[width=4.0cm,height=2.7cm]{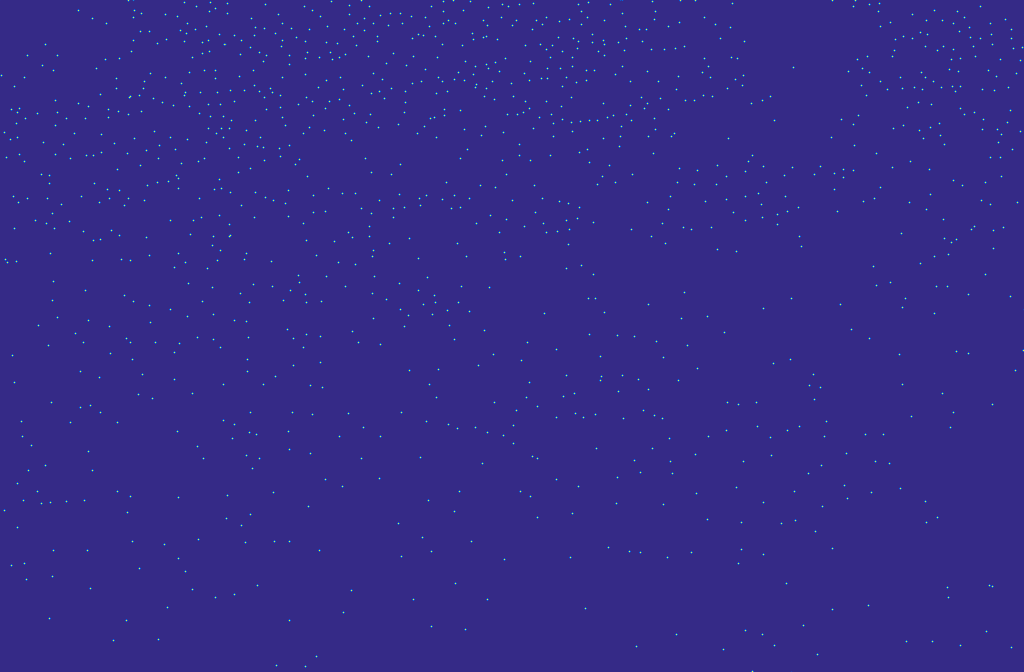}};
    \node[below=of img3, anchor=center, yshift=0.85cm, ] {(c) Ours: 1168};

    \node[right=of img3,xshift=-1.2cm] (img4) 
    {\includegraphics[width=4.0cm,height=2.7cm]{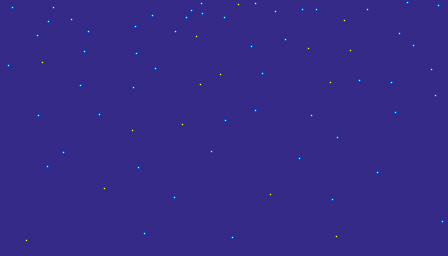}};
    \node[below=of img4, anchor=center, yshift=0.85cm, ] {(d) Ours: 69};

    \node[below=of img3, yshift=0.75cm] (img5) 
    {\includegraphics[width=4.0cm,height=2.7cm]{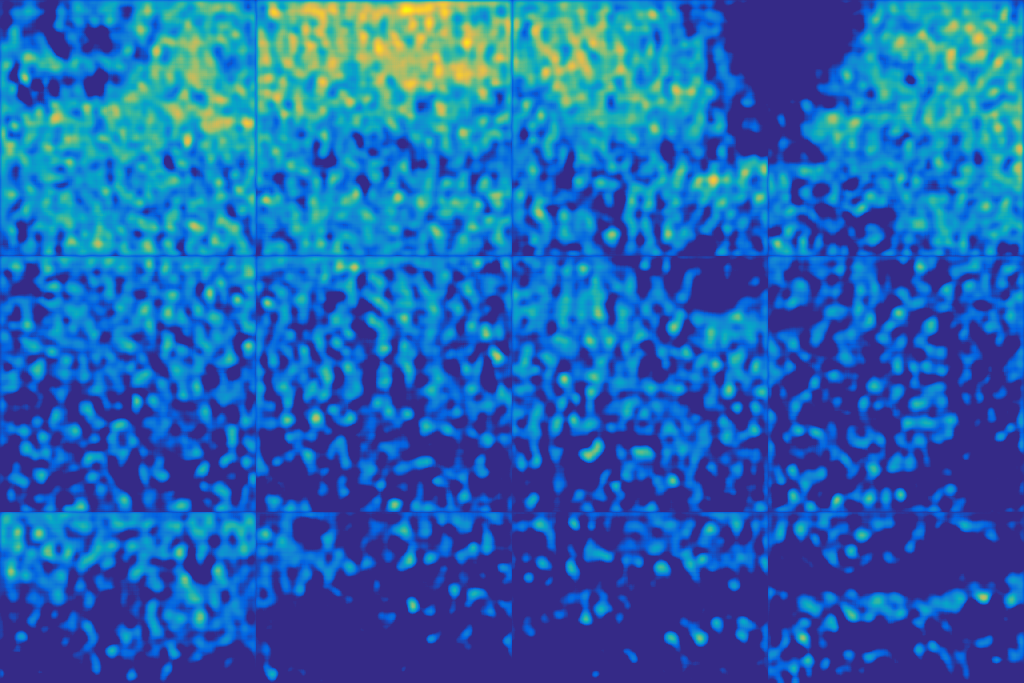}};
    \node[below=of img5, anchor=center, yshift=0.85cm, ] {(c) ACSCP: 1765.36};

    \node[right=of img5,xshift=-1.2cm] (img6) 
    {\includegraphics[width=4.0cm,height=2.7cm]{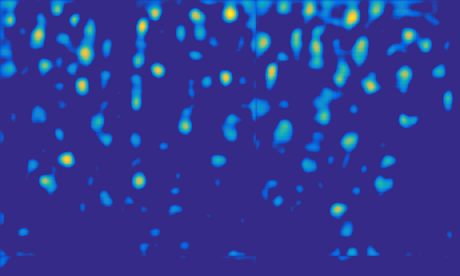}};
    \node[below=of img6, anchor=center, yshift=0.85cm, ] {(d) ACSCP: 52.00};
    
    \node[right=of img3,xshift=-3.375cm,yshift=1.08cm] (frame1) 
    {\includegraphics[height=0.10cm]{figures/intro_crops/box.pdf}};

    \node[right=of img3,xshift=-2.235cm,yshift=-0.865cm] (frame1) 
    {\includegraphics[height=1.0cm]{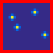}};
    
\end{tikzpicture}

    \caption{Generation quality and crowd performance comparison with a narrow kernel between the diffusion models and a GAN-based method (ACSCP) \cite{shen2018crowd} for different crowd scenes.}
    
    \label{figure: gan ablation}
    
\end{figure}


\subsection{Ablation study}
\label{subsec: ablations}

\noindent\textbf{Diffusion models} are considered to have more fidelity to training data than GAN-based methods. From \cref{figure: gan ablation}, one can see that diffusion models have produced high-quality density maps with more accurate crowd counts while ASCSP \cite{shen2018crowd}, a GAN-based method, has failed. Furthermore, without the ability to produce narrow kernels in the predicted density maps, GAN-based methods have to use density summation as the counting operation, bringing back noise accumulation and loss of density. This highlights the importance of using diffusion models for crowd density map generation with detection as the counting operation.

\begin{figure}[!t]
    \centering
\begin{tikzpicture}
    \node (img1)
    {\includegraphics[height=2.8cm, width=3.8cm]{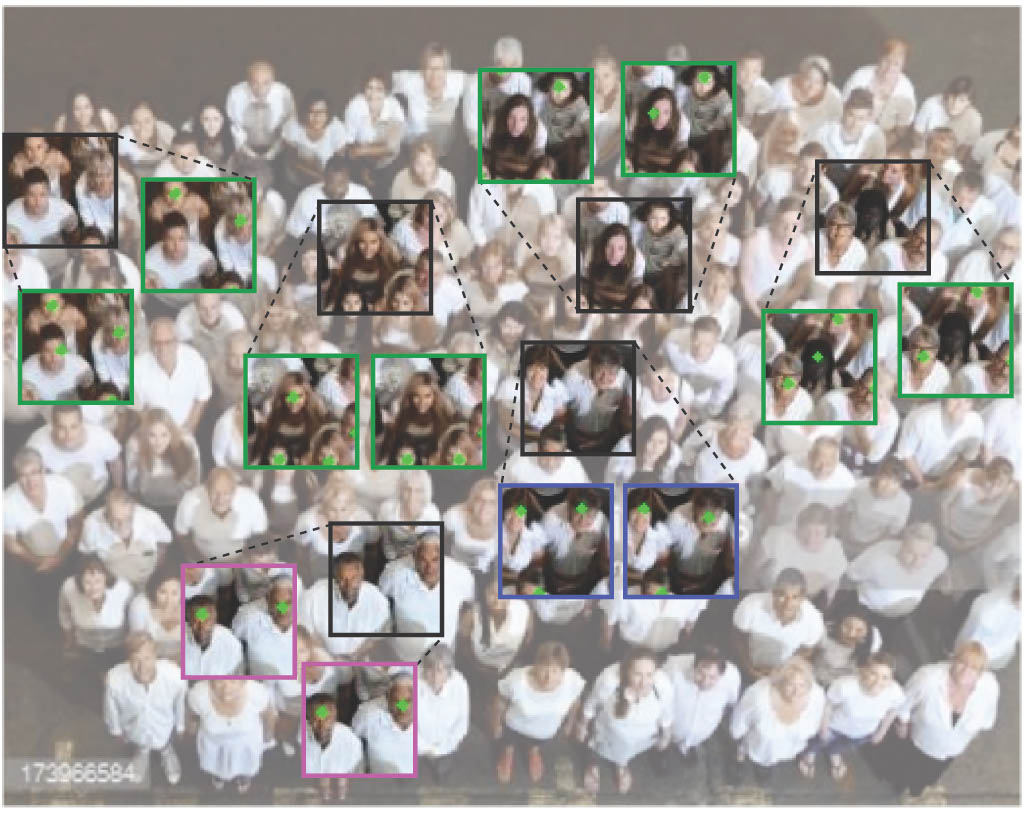}};    
    \node[right=of img1,xshift=-1.2cm] (img2) 
    {\includegraphics[height=2.8cm]{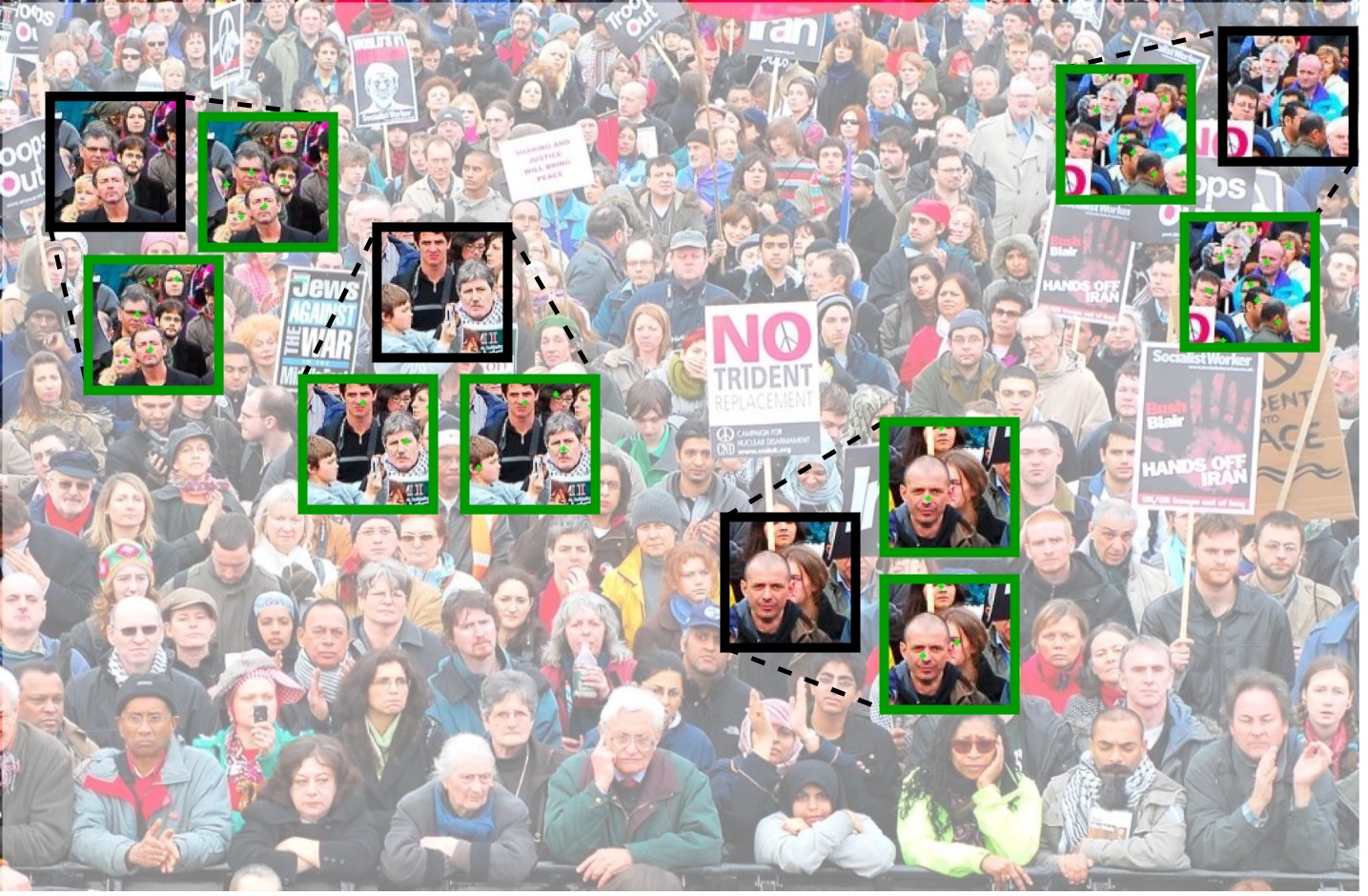}};
    \node[below=of img1, yshift=1.22cm, xshift=2.13cm] (img5) 
    {\includegraphics[width=0.977\columnwidth]{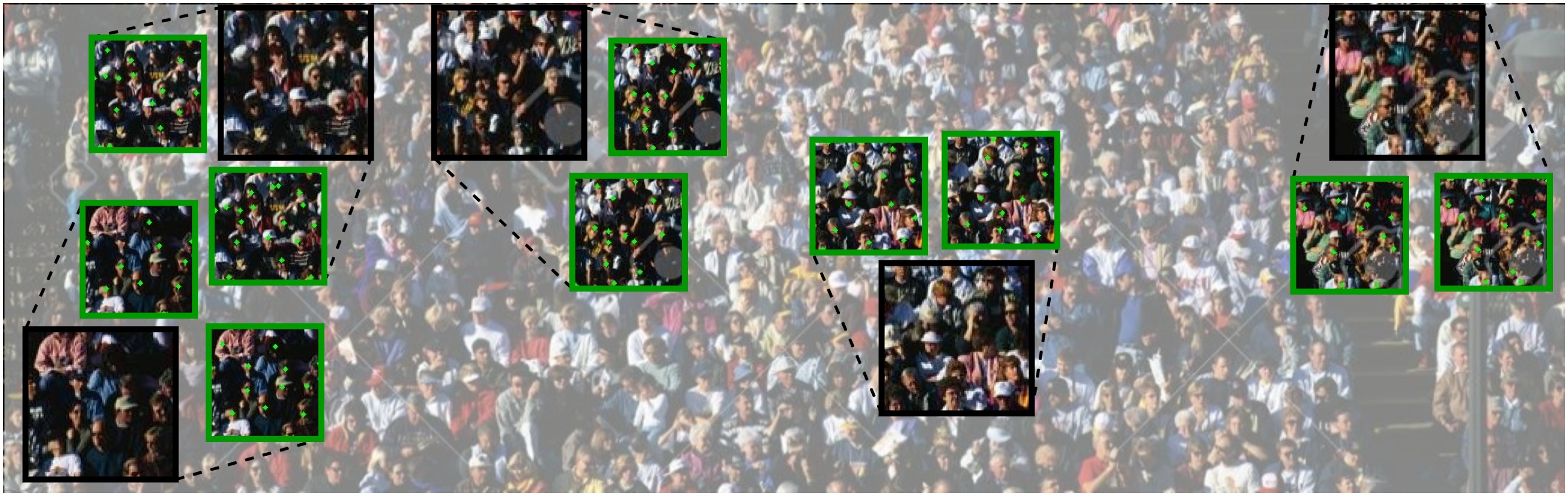}};   
\end{tikzpicture}

    \caption{Qualitative results for stochastic crowd map generation from two realizations. Green boxes include new dots created at different realizations. Blue boxes include dots present in both realizations but with a shift, and pink boxes include perfectly reassigned dots. ({\it best viewed in highest zooming level})}
    
    \label{figure: stochastic image generation}
    
\end{figure}

\noindent\textbf{Stochastic crowd map} generation is a key benefit of diffusion-based generative models. In \cref{figure: stochastic image generation}, we provide qualitative results of two realizations for each crowd image. From  \cref{figure: stochastic image generation}, we can see that different realizations have information that is not present in other realizations. Furthermore, it is noteworthy that using a narrow kernel facilitates the ability to produce new knowledge that can be included in the final prediction. Otherwise, novel information captured by different realizations will be diluted by averaging the density maps had a larger kernel been used. Though this is a generative model, the proposed method has reassigned densities perfectly in certain instances, and for some cases, there is a slight shift in the location between realizations. This necessitates the need for the proposed crowd map fusion method since simply combing these shifted dots results in double counts and worsens the  performance otherwise.\\
\textbf{The counting branch} is added to improve the counting performance with the crowd density maps. We present the counting results for individual realizations with and without the regression network in \cref{table: counting branch ablation}. We consider the average error performance from different realizations to identify characteristic effects of the counting branch before combining them into a single crowd map. Adding the counting branch has improved the average counting results, and the variation in the counting results has been reduced. The counting branch also promotes feature learning for intermediate time steps with noisy features.

Furthermore, we considered the performance of the counting branch even though it is not used to predict the final count of \method. The error metrics for the counting branch are provided in \cref{table: counting performance ablation} along with \sota\ weakly-supervised crowd counting methods \cite{lei2021towards, liang2022transcrowd}. The counting branch can be considered as a subnetwork that was weakly supervised with features from the denoising network and in this regard, the counting branch of \method\ outperforms existing SOTA weakly-supervised methods.

\begin{figure*}[!t]
\begin{center}

   \includegraphics[width=\linewidth]{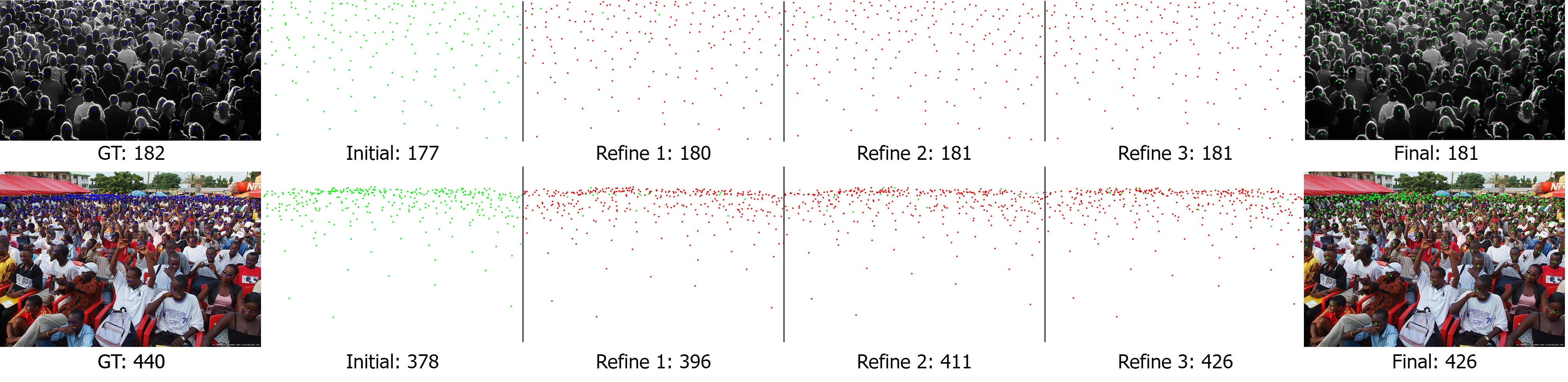}
\end{center}

  \caption{Qualitative results for the crowd density map fusion method. Green dots represent the points combined to the final prediction and red dots represent the points removed from each realization.}
  
\label{figure: density fusion ablation}
\end{figure*}

\noindent\textbf{Crowd map fusion} leverages the stochastic nature of the crowd density maps produced by the diffusion process, and we adopt a systematic way to fuse the maps. In \cref{table: fusion method ablation}, we present the error metrics for three different methods: Random, Descend-SSIM, and Ascend-SSIM. In the {\it Random} method, we combine the maps in the order in which they are produced. In the {\it Descend-SSIM} method, we combine the maps in the order of decreasing similarity. In the {\it Ascend-SSIM} method, we combine the maps in the order of increasing similarity as described above. The iterative improvement with stochastic generation and the proposed fusion method is shown in \cref{figure: density fusion ablation}. From  \cref{table: fusion method ablation}, the counting performance has improved with the Ascend method, where more locally dissimilar realizations are combined initially. This observation is validated by the performance degradation with the Descend-SSIM method compared to both Ascend-SSIM and Random methods.

Additionally, the fusion of multiple realizations is prone to introduce false positives. Hence, we considered the localization and counting performance after fusing different realizations for \qnrf. We generated four additional realizations for this ablation study, and the corresponding results are provided in \cref{table: localization ablation} along with the respective inference time. From \cref{table: localization ablation}, we see that the localization and counting performance improves with multiple realizations, demonstrating the advantage of using a generative model and fusing the information from multiple realizations. However, a higher number of realizations increases the inference time, and the performance gain from four realizations to eight realizations is insignificant while the inference time has doubled. Consequently, we chose to produce four realizations as the optimal setting considering the performance and inference time trade-off.\\
\textbf{Density thresholding} is used as an alternative to density summation for the counting operation. The performance comparison between the two methods is tabulated in \cref{table: counting method ablation} for the best-performing realization of each dataset. From  \cref{table: counting method ablation}, we see that the density summation produces  inferior counting results than thresholding despite both methods using the same density map. This is because background noise accumulation with the summation operation and the thresholding method display better noise immunity.

\noindent\textbf{Kernel size} of the density kernels used to generate the ground truth density map $\mathbf{x_0}$ affects the generation ability and performance of \method. We tabulate the performance with different kernel sizes in \cref{table: kernel ablation}.
We observed similar performance at $1 \times 1$ and $3 \times 3$ kernels, and the performance significantly dropped for the latter kernel sizes. This is because the kernel size affects the pixel value distribution of the density map, and the interference between adjacent kernels introduces true positives at local maxima.\\
\textbf{Rejection radius ($\beta$) and nearest neighbors ($k$)} influence the performance of the fusion of multiple realizations. The results for different $\beta$ and $k$ values are tabulated in \cref{table: beta ablation}. The rejection criterion is stable around the $\beta$ values from $0.80$ to $0.85$. Because a low $\beta$ value is susceptible to including false positives or duplicates, and a high $\beta$ value could also reject true positives. However, the performance difference for different $k$ values was insignificant and the best performing setting was selected.\\
\textbf{The inference process} of the diffusion-based models is iterative and, therefore, exacts higher inference times. However, since we threshold the density map to count the number of kernels rather than summing over the pixel density values, the proposed method is robust to residual noise in the background. With the above exception, we used DDIM sampling to improve the inference procedure by a factor of 20 rather than using the original number of diffusion steps without a significant performance drop. \\
More results and details can be found in the supplementary.

\section{Conclusions}
We proposed a novel crowd counting framework where density map generation was treated as a denoising diffusion process. The new framework allows using extremely narrow density kernels with which noise can be suppressed more robustly in crowd density maps. Consequently, we performed density kernel detection on crowd density maps which offered more immunity to noise than density summation. Also, the proposed method could iteratively improve the counting performance via multiple realizations, unlike other crowd counting frameworks, due to the stochastic nature of the generative models. Further, unlike existing density-based methods, our proposed method assigns density kernels at head positions without the need for data heuristics, as required in the localization-based methods.

\begin{table}[!t]
\begin{center}
\caption{Error metrics for individual realizations without (top half) and with (bottom half) the counting decoder.}
\vspace{\tablegap}
\resizebox{\linewidth}{!}{
\begin{tabular}{@{}l l c c c c c c@{}}
\toprule
\multirow{2}{*}{~} & \multirow{2}{*}{Method} &\multicolumn{2}{c}{\jhu}  &\multicolumn{2}{c}{\shhb} &\multicolumn{2}{c}{\qnrf}\\
\cmidrule(lr){3-4}\cmidrule(lr){5-6}\cmidrule(lr){7-8}
&~ & MAE$\downarrow$ & MSE$\downarrow$ & MAE$\downarrow$ & MSE$\downarrow$ & MAE$\downarrow$ & MSE$\downarrow$ \\[0.2ex]
\midrule\midrule
\multirow{2}{*}{\rotatebox[origin=c]{90}{\small w/o}}
& Best  & 50.24	    & 206.82	& 5.90	    & 8.40	  & 75.87	& 136.85	\\[0.0ex]
& Average   & 52.29	    & 212.22	& 5.97	    & 8.50	  & 78.35	& 140.87	\\[0.0ex]
& Variance  & 1.5854	& 4.1764	& 0.0926	& 0.1278  & 2.3092	& 3.7404	\\[1ex]
\multirow{2}{*}{\rotatebox[origin=c]{90}{\small w/}}
& Best  & 48.24	    & 201.54	& 5.82	    & 8.30	  & 72.17	& 130.86	\\[0.0ex]
& Average   & 48.56	    & 202.38	& 5.85	    & 8.33	  & 73.08	& 132.33	\\[0.0ex]
& Variance  & 0.2546	& 0.6708	& 0.0209	& 0.0289  & 0.64	& 1.0366	\\[0.1ex]
\bottomrule
\end{tabular}
}
\vspace{\tablegap}
\label{table: counting branch ablation}
\end{center}
\end{table}

\begin{table}[!t]
\begin{center}
\caption{Performance of the counting branch in comparison to other weakly-supervised counting methods.}
\vspace{\tablegap}
\resizebox{\linewidth}{!}{
\begin{tabular}{@{}l c c c c c c@{}}
\toprule
\multirow{2}{*}{Method} &\multicolumn{2}{c}{\jhu}  &\multicolumn{2}{c}{\shhb} &\multicolumn{2}{c}{\qnrf}\\
\cmidrule(lr){2-3}\cmidrule(lr){4-5}\cmidrule(lr){6-7}
& MAE$\downarrow$ & MSE$\downarrow$ & MAE$\downarrow$ & MSE$\downarrow$ & MAE$\downarrow$ & MSE$\downarrow$ \\[0.2ex]
\midrule\midrule
Counting	& 53.1	& 223.5 & 7.7	& 12.0	& 76.6	& 135.3\\
TransCrowd \cite{liang2022transcrowd}	& 56.8	& 193.6 & 9.3	& 16.1	& 97.2	& 168.5\\
MATT \cite{lei2021towards}	    & 71.5	& 210.4 & 11.7	& 17.5	& 122.3	& 183.2\\
\bottomrule
\end{tabular}
}
\vspace{\tablegap}
\label{table: counting performance ablation}
\end{center}
\end{table}

\begin{table}[!t]
\begin{center}
\caption{Comparison for crowd map fusion methods.}
\vspace{\tablegap}
\resizebox{\linewidth}{!}{
\begin{tabular}{@{}l c c c c c c@{}}
\toprule
\multirow{2}{*}{Method} &\multicolumn{2}{c}{\jhu} &\multicolumn{2}{c}{\shhb} &\multicolumn{2}{c}{\qnrf}\\
\cmidrule(lr){2-3}\cmidrule(lr){4-5}\cmidrule(lr){6-7}
& MAE$\downarrow$ & MSE$\downarrow$ & MAE$\downarrow$ & MSE$\downarrow$ & MAE$\downarrow$ & MSE$\downarrow$ \\[0.2ex]
\midrule\midrule
Random & 47.77	& 200.3	& 5.78	& 8.23	& 71.04	& 129.03 \\[0.0ex]
Ascend-SSIM & 47.26	& 198.97	& 5.74	& 8.18	& 68.95	& 125.65 \\[0.0ex]
Descend-SSIM & 48.10	& 201.18	& 5.81	& 8.27	& 71.73	& 130.15 \\[0.1ex]
\bottomrule
\end{tabular}
}
\vspace{\tablegap}
\label{table: fusion method ablation}
\end{center}
\end{table}

{\small
\section*{Acknowledgements}
This research is based upon work supported in part by the Office of the Director of National Intelligence (ODNI), Intelligence Advanced Research Projects Activity (IARPA), via [2022-21102100005]. The views and conclusions contained herein are those of the authors and should not be interpreted as necessarily representing the official policies, either expressed or implied, of ODNI, IARPA, or the U.S. Government. The US Government is authorized to reproduce and distribute reprints for governmental purposes notwithstanding any copyright annotation therein.
}

\begin{table}[!t]
\tiny
\begin{center}
\caption{Counting and localization results with the number of realizations.}
\vspace{\tablegap}
\resizebox{\linewidth}{!}{
\begin{tabular}{@{}ccccccc@{}}
 \toprule
 \multirow{2}{*}{\#} &\multirow{2}{*}{\parbox{5mm}{Time\\(ms)}} &\multicolumn{2}{c}{Counting} &\multicolumn{3}{c}{Localization}  \\
\cmidrule(lr){3-4}\cmidrule(lr){5-7}
& &MAE$\downarrow$ &MSE$\downarrow$ & P (\%)$\uparrow$& R (\%)$\uparrow$ &F (\%)$\uparrow$  \\[0.2ex]
\midrule\midrule
1 & 210 & 74.59	& 134.78 & 68.45 & 67.34 & 67.89\\
2 & 430 & 71.94	& 130.49 & 77.24 & 75.94 & 76.58\\
4 & 770 & 68.95	& 125.65 & 82.18 & 80.79 & 81.48\\
8 & 1360 & 66.97	& 122.44 & 83.06 & 81.67 & 82.36\\
 \bottomrule
\end{tabular}
}
\vspace{\tablegap}
\label{table: localization ablation}
\end{center}
\end{table}


\begin{table}[!t]
\centering
\caption{Performance comparison for different kernel sizes.}
\vspace{\tablegap}
\setlength{\tabcolsep}{1mm}
\resizebox{0.8\linewidth}{!}{
\begin{tabular}{@{}ccccccc@{}}
    \toprule
    Kernel size & $\sigma$   & MAE$\downarrow$ & MSE$\downarrow$ & P(\%)$\uparrow$ & R(\%)$\uparrow$ & F(\%)$\uparrow$\\
    \midrule\midrule
    $1\times 1$	& - & 69.21	& 126.07 & 81.70 & 80.32 & 81.00\\
    $3\times 3$	& 0.5 & 68.95	& 125.65 & 82.18 & 80.79 & 81.48\\
    $5\times 5$	& 1 & 81.87	& 146.58 & 57.43 & 56.52 & 56.97\\
    $9\times 9$	& 2 & 94.68	& 167.32 & 34.12 & 33.65 & 33.88\\
\bottomrule
\end{tabular}
}
\vspace{\tablegap}
\label{table: kernel ablation}
\end{table}

\begin{table}[!t]
\begin{center}
\caption{Comparison between crowd counting operations and the effect of noise.}
\vspace{\tablegap}
\resizebox{\linewidth}{!}{
\begin{tabular}{@{}l c c c c c c@{}}
\toprule
\multirow{2}{*}{Method} &\multicolumn{2}{c}{\jhu} &\multicolumn{2}{c}{\shha} &\multicolumn{2}{c}{\ucf}\\
\cmidrule(lr){2-3}\cmidrule(lr){4-5}\cmidrule(lr){6-7}
& MAE$\downarrow$ & MSE$\downarrow$ & MAE$\downarrow$ & MSE$\downarrow$ & MAE$\downarrow$ & MSE$\downarrow$ \\[0.2ex]
\midrule\midrule
Density thresholding	& 48.24	    & 201.54	& 47.81	   & 75.91	& 163.56 & 228.32	\\[0.0ex]
Density estimation	& 215.40	& 515.63	& 156.96   & 243.70	& 180.58 & 254.91	\\[0.0ex]
Noise residual	& 200.94	& 502.31	& 186.04   & 294.44	& 70.68	 & 99.38	\\[0.1ex]
\bottomrule
\end{tabular}
}	
\vspace{\tablegap}
\label{table: counting method ablation}
\end{center}
\end{table}

\begin{table}[!t]
\centering
\caption{Ablation for different rejection radii ($\beta$) and nearest neighbor ($k$) values.}
\vspace{\tablegap}
\resizebox{\linewidth}{!}{
\begin{tabular}{@{}lccccccc@{}}
    \toprule
    \multirow{3}{*}{Metric} & \multicolumn{4}{c}{Rejection radius ($\beta$)} & \multicolumn{3}{c}{Nearest neighbors ($k$)}\\
    & \multicolumn{4}{c}{$k=4$} & \multicolumn{3}{c}{$\beta=0.85$}\\
    \cmidrule(lr){2-5}\cmidrule(lr){6-8}
    & 0.75 & 0.80 & 0.85 & 0.90 & 3 & 4 & 5\\
    \midrule\midrule
    MAE$\downarrow$ & 72.2	& 69.71	& 68.95	& 72.03	& 69.41	& 68.95	& 69.07 \\
    MSE$\downarrow$ & 130.91 & 126.88 & 125.65 & 130.64 & 126.39 & 125.65 & 125.84 \\
    P(\%)$\uparrow$ & 79.58	& 81.67	& 82.18	& 81.64	& 81.83	& 82.18	& 82.01 \\
    R(\%)$\uparrow$ & 78.68	& 79.94	& 80.79	& 79.36	& 80.40	& 80.79	& 80.55 \\
    F(\%)$\uparrow$ & 79.13	& 80.80	& 81.48	& 80.48	& 81.11	& 81.48	& 81.27 \\
    \bottomrule
\end{tabular}
}
\vspace{\tablegap}
\label{table: beta ablation}
\end{table}


\clearpage

\setcounter{section}{0}
\maketitlesupplementary
\renewcommand{\thesection}{\Alph{section}}

\numberwithin{figure}{section}
\numberwithin{table}{section}

\renewcommand{\thefigure}{\thesection.\arabic{figure}}
\renewcommand{\thetable}{\thesection.\arabic{table}}

\makeatletter
\renewcommand{\thealgorithm}{\thesection.\arabic{algorithm}}  
\@addtoreset{algorithm}{section}
\makeatother

\section{Pseudocodes}

The pseudocode for training is given in \cref{alg: train code} and testing in \cref{alg: test code} for \method.
\vspace{-1mm}
\begin{algorithm}[!ht]
\caption{Training phase}
\label{alg: train code}
\begin{lstlisting}[language=python]
def train(images,density_maps,gt_counts):
    """
        images: [B, H, W, 3]
        density_maps: [B, H, W]
        gt_counts: [B,]
    """
    
    # Density scaling
    density_maps = (2*scale*density_maps-1)

    # Corrupt density_maps
    t = randint(0,T) # time step
    eps = normal(mean=0,std=1)  # noise: [B,H,W]
    crpt_density_maps = 
      diffusion_process(density_maps,eps,t)
    
    # Estimate noise and encoder-decoder features
    eps_pred, feats =
      denoising_network(images,crpt_density_maps,t)

    # Estimate crowd count
    count_est = counting_decoder(feats)
       
    # Compute denoising network loss
    loss = 
      l_hybrid(eps_pred, eps) + 
      count_scale * l1_loss(count_est,gt_count)
    
    return loss
\end{lstlisting}

\end{algorithm}
\vspace{-2.5mm}


\begin{algorithm}[!b]
\caption{Testing phase}
\label{alg: test code}
\begin{lstlisting}[language=python]
def testing(images, realizations):
    """
    images: [B, H, W, 3]
    realizations: N
    """
    
    # Encode image features
    feats = image_encoder(images)
    
    # noisy density maps: [B, H, W]
    density_pred = normal(mean=0, std=1)
    
    # uniform sample step size
    times = reversed(
    linespace(diffusion_steps, sampling_steps))

    # Perform DDIM sampling
    for t in times:
        # Predict noise from density_pred
        eps_hat = denoising_network(images, noisy_density, t)
        # Compute posterior of noisy density
        density_pred = q_posterior(noisy_density, eps, t)

    # Detect head locations
    locations = contours(density_pred) # [B, N, *, 2]

    # Perform crowd map fusion: [B, *, 2]
    final_locations = crowd_map_fusion(locations)

  # Compute crowd count    
  return count(final_locations) # [B, ]
\end{lstlisting}

\end{algorithm}
\newpage

\section{Experimental details}
\begin{figure*}[!ht]
\begin{center}
   \includegraphics[width=1.0\linewidth]{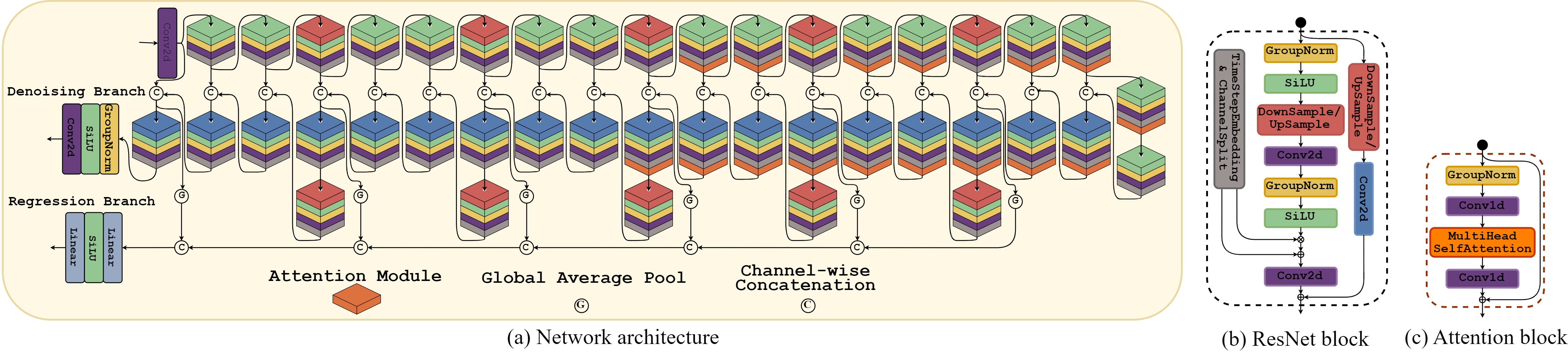}
\end{center}
    
    \caption{(a) {\bf Network architecture} for the denoising U-Net in conjunction with the count regression branch and the basic modules, (b) {\bf ResNet block}, and (c) {\bf Attention module}, used to construct the network. Each cuboid in a stack represents the functioning modules in the ResNet Block and whether the attention module is applied. {\bf Top stacks} are in the encoder. {\bf Bottom stacks} are in the decoder.}
    \label{figure: network architecture}
    
\end{figure*}

1. Denoising network architecture
\noindent {\bf{Denoising network}} has a U-Net architecture \cite{nichol2021improved}, and each downsampling and upsampling layer scales the features by a factor of two along each spatial dimension. We use average pooling for downsampling with a $2\times2$ kernel, a stride of 2, and nearest neighbor interpolation for upsampling. The 2-dimensional convolution layers are $3\times3$ kernels with a stride of 1, and the 1-dimensional convolution layers have a kernel size and a stride of 1. In the multi-head self-attention module, the channel dimension of each head is kept constant at 64, and the number of heads is varied according to the channel dimension of each depth level. The denoising network and the basic modules are illustrated in \cref{figure: network architecture}.

\noindent {\bf{Regression branch}} is a lightweight network with linear layers and a Rectified Linear Unit (ReLU) \cite{agarap2018deep} activation. We apply global average pooling to maintain compatibility along the spatial dimension for channel-wise concatenation.

\section{Evaluation metrics}
To evaluate crowd counting performance, we use the mean absolute error (MAE):
\[
    MAE = \frac{1}{N}\sum_{n=1}^{N} {\lVert c_n - \bar{c_n} \rVert}_1,
\]
and root mean squared error (MSE):
\[
    MSE = \sqrt{\frac{1}{N}\sum_{n=1}^{N} {\lVert c_n - \bar{c_n} \rVert}_2^2}
\]
as the performance metrics. Here, $N$ is the total number of test samples, $c_n$ is the ground truth count, and $\bar{c_n}$ is the prediction for the $n$\textsuperscript{th} sample. 

\section{Datasets}

We evaluate our method on five public datasets: \jhu \cite{sindagi2019pushing}, \shha \cite{zhang2016single}, \shhb \cite{zhang2016single}, \ucf \cite{idrees2013multi}, and \qnrf \cite{idrees2018composition} for crowd counting.\\
{\bf \jhu} \cite{sindagi2019pushing} has 2,722 training images, 500 validation images, and 1,600 test images collected from diverse scenarios. The dataset consists of crowd images with numbers ranging up to 25,791 and images without any crowd.\\
{\bf \shha} \cite{zhang2016single} contains 300 training images and 182 test images with annotations. We randomly select 30 samples from the training dataset as the validation dataset.\\
{\bf \shhb} \cite{zhang2016single} contains 400 training images and 316 testing images with annotations. We create a validation dataset with randomly selected 40 crowd images from the training dataset.\\
{\bf \ucf} \cite{idrees2013multi} is a comparatively small crowd dataset for extremely dense crowd counting with just 50 samples. We perform a 5-fold cross-validation following the standard protocol in \cite{idrees2013multi}.\\
{\bf \qnrf} \cite{idrees2018composition} dataset contains 1,535 images of unconstrained crowd scenes, with approximately one million annotations in total. The dataset is split into a training set of 1,201 images and a testing set of 334 images.\\
\textbf{\nwpu} 
NWPU-Crowd \cite{wang2020nwpu} is a large-scale dataset collected from various scenes, consisting of 5,109 images. The images are randomly split into training, validation, and test sets containing 3109, 500, and 1500 images, respectively. This dataset provides box-level annotations.

\section{Additional qualitative results}
\begin{figure*}[!t]
    \centering
    \includegraphics[width=1\linewidth]{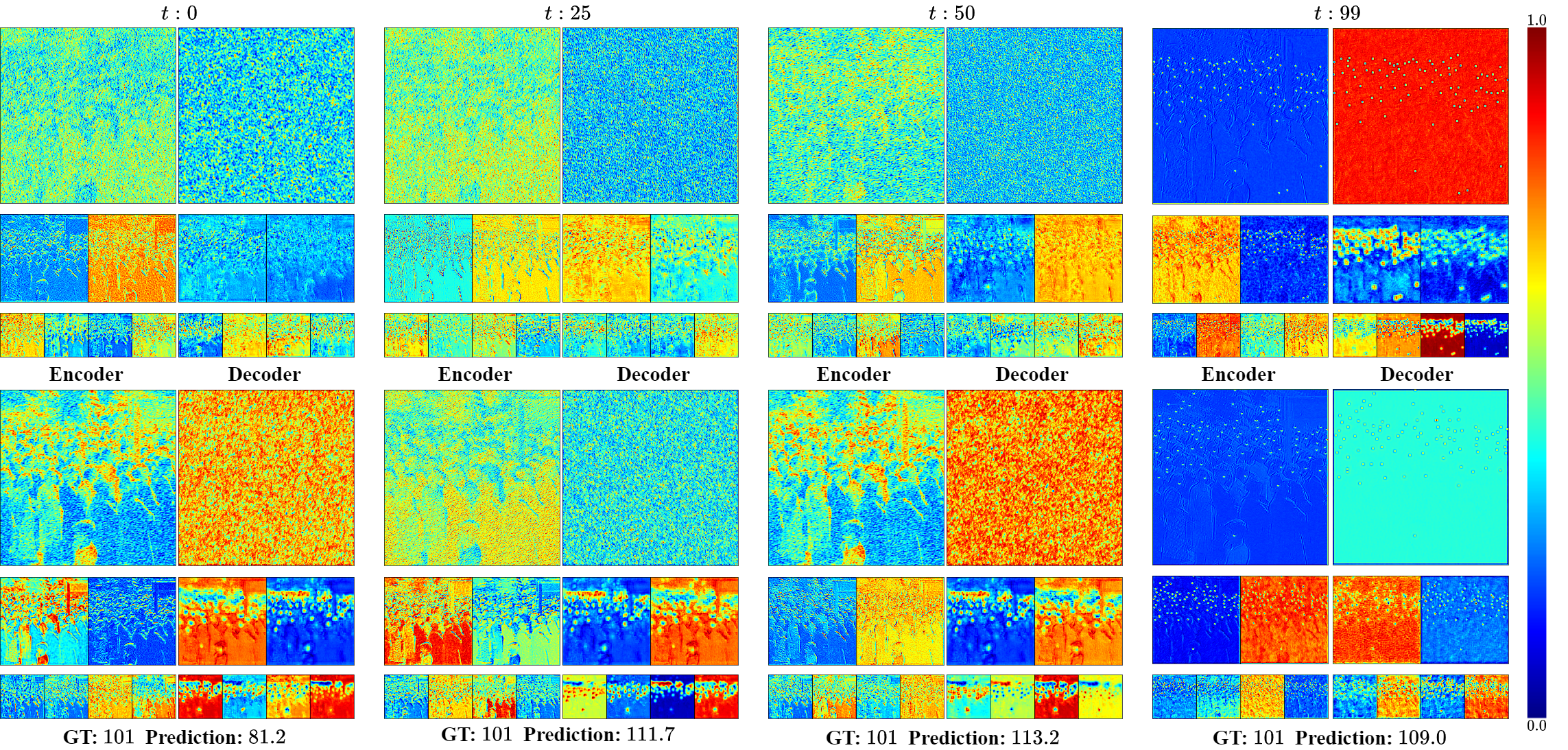}
    \caption{Difference in feature maps without (\textit{top row}) and with (\textit{bottom row}) counting decoder.}
    \label{figure: feature maps}
\end{figure*}

We provide a qualitative comparison between the feature maps of the denoising U-Net with and without the counting branch prediction for different time steps in \cref{figure: feature maps}. From \cref{figure: feature maps}, we can see that the decoder features are richer in detail for the case with the counting branch than without it. With the counting branch, the decoder generates features for the crowd starting from the initial time step. The performance of the counting branch further clarifies this, as the predicted count has not varied with time and deviated from the ground truth count significantly.

\section{Number of inference steps}
\begin{figure}[!ht]
    \centering
    \includegraphics[width=0.8\linewidth]{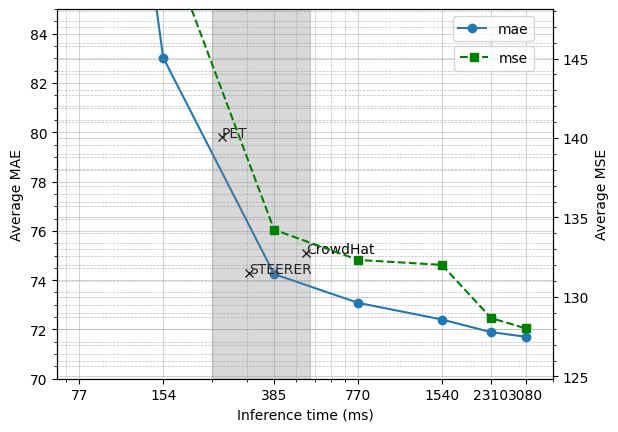}
    
    \caption{Performance variation with sampling steps}
    \label{figure: steps vs metrics}
\end{figure}

Though the diffusion model was trained using 1,000 diffusion steps, we can perform inference with fewer steps using DDIM sampling \cite{song2020denoising}. However, selecting the number of sampling steps with a good compromise between the inference speed and MAE performance is pertinent. We considered different sampling steps and the corresponding inference speed and performance to decide the number of inference steps. We then compared it with the inference speed and performance of state-of-the-art methods to find an optimal number of sampling steps. We display the variation between performance and inference speed in \cref{figure: steps vs metrics} for MAE and MSE with \method. In \cref{figure: steps vs metrics}, the values are provided for the average from four realizations on \qnrf\ benchmark, and the shaded region marks the interval of the inference speed for the most recent crowd analysis methods: PET \cite{liu2023point}, STEERER \cite{han2023steerer}, and CrowdHat \cite{wu2023boosting}. Besides that, recently published ``consistency models" \cite{song2023consistency} could improve the sampling quality of few-step inference and facilitate single-step inference.

\section{Training setting}
The training parameters used for the denoising network and the counting decoder are presented in \tab{\ref{table: training setting}}.

\begin{table}[!ht]
\caption{Training parameters for the crowd counting network}
\label{table: training setting}

\begin{center}
\tiny
\resizebox{0.5\linewidth}{!}{
\begin{tabular}{l c}
\hline \\[-2ex]
Configuration & setting \\[0.2ex]
\hline \\[-2ex]
Optimizer & Adamw\\
Optimizer betas & \{0.9, 0.999\}\\
Base learning rate & 1e-4\\
Warmup steps & 5000\\
Training steps & 2e5\\
Image size & 256$\times$256\\
Batch size &8\\
Diffusion steps & 1000\\
Noise schedule & Linear\\

\hline
\end{tabular}
}
\end{center}
\end{table}

{
    \small
    \bibliographystyle{ieeenat_fullname}
    \bibliography{main}
}

\end{document}